\documentclass{article} 
\usepackage{iclr2027_conference,times}

\usepackage{enumitem}

\usepackage{amsmath,amsfonts,bm}

\def\eqref#1{equation~\ref{#1}}

\def\1{\bm{1}}

\DeclareMathAlphabet{\mathsfit}{\encodingdefault}{\sfdefault}{m}{sl}
\SetMathAlphabet{\mathsfit}{bold}{\encodingdefault}{\sfdefault}{bx}{n}

\usepackage{hyperref}
\usepackage{url}

\usepackage{amsmath}
\usepackage{amsfonts}
\usepackage{booktabs}
\usepackage{xcolor}
\usepackage{algorithm}      
\usepackage{algpseudocode}  
\usepackage{graphicx}
\usepackage{float}          
\usepackage{placeins}       
\usepackage{comment}        
\usepackage[font=small,labelfont=bf,skip=5pt]{caption}  

\def\paperpart{full}
\makeatletter
\def\@partmain{main}\def\@partappendix{appendix}
\newif\if@droppages
\AddToHook{shipout/before}{\if@droppages\DiscardShipoutBox\fi}
\def\@startdropping{%
  \global\@droppagestrue\global\@fileswfalse
  \let\@keptlabel\label\let\@keptltxlabel\ltx@label
  \def\label##1{{\let\protected@write\@immediatewrite\@keptlabel{##1}}}%
  \def\ltx@label##1{{\let\protected@write\@immediatewrite\@keptltxlabel{##1}}}}
\long\def\@immediatewrite#1#2#3{
  \begingroup#2\let\protect\@unexpandable@protect
  \edef\reserved@a{\immediate\write#1{#3}}\reserved@a\endgroup}
\def\@stopdropping{%
  \global\@droppagesfalse\global\@fileswtrue
  \let\label\@keptlabel\let\ltx@label\@keptltxlabel}
\AddToHook{enddocument}{\global\@fileswtrue}
\newcommand{\beginmainpart}{\ifx\paperpart\@partappendix\@startdropping\fi}
\newcommand{\beginappendixpart}{%
  \cleardoublepage
  \ifx\paperpart\@partmain\@startdropping\fi
  \ifx\paperpart\@partappendix
    \@stopdropping\setcounter{page}{1}%
    {\centering\Large Supplementary material for\\[4pt]\@savedtitle\par}\vspace{1em}%
  \fi}
\newcommand{\mainreferences}{\ifx\paperpart\@partappendix\else\paperreferences\fi}
\newcommand{\appendixreferences}{\ifx\paperpart\@partappendix\paperreferences\fi}
\makeatother
\newcommand{\paperreferences}{%
  \bibliography{iclr2027_conference}%
  \bibliographystyle{iclr2027_conference}}

\title{Continual Reinforcement Learning with \\Neuroevolution}
\makeatletter\let\@savedtitle\@title\makeatother  

\author{Eleni Nisioti$^{1}$, Andrea Cossu$^{2}$, Kathrin Korte$^{1}$, Sebastian Risi$^{1,3}$ \\
$^{1}$IT University of Copenhagen \quad $^2$University of Pisa \quad $^{3}$Sakana AI}

\iclrfinalcopy 
\begin{document}
\beginmainpart

\maketitle
\lhead{Preprint}
\addtocontents{toc}{\protect\setcounter{tocdepth}{-1}}

\begin{abstract}
Despite many studies about causes and remedies of plasticity loss in Reinforcement Learning (RL) under continual task changes, no RL method has yet consistently achieved a good balance between adaptation and forgetting.
Here we turn to an alternative optimization paradigm, neuroevolution (NE): algorithms that search directly in weight space through mutation and selection over a population of neural networks.
Across a wide array of environments and environmental changes, with policies ranging from a few hundred parameters to million-parameter networks, we compare evolution strategies (ES) and genetic algorithms (GAs) against state-of-the-art continual RL variants and population-based RL.
ES most consistently achieves a good stability-plasticity trade-off, while the GA is the most plastic method but forgets more than ES.
To explain this, we study the return landscape around each method's solutions.
ES finds the widest neighborhoods, i.e.\ regions of weight space in which perturbed policies still solve the task, and the size of the overlap between the neighborhoods of consecutive tasks correlates with a method's stability-plasticity trade-off.
Rewarding behavioral diversity in a GA through novelty search makes the population even more plastic, at the cost of forgetting.
Finally, symptoms of plasticity loss commonly reported in RL do not transfer to NE.
Overall, these results establish NE as a competitive alternative to RL under continual task changes, and suggest that training under perturbations in weight space may be a useful mechanism for continual learning more broadly.
\end{abstract}

\section{Introduction}

Reinforcement learning (RL) has achieved remarkable successes, ranging from superhuman gameplay \citep{silver2016mastering} to the control of high-dimensional robots \citep{DBLP:journals/corr/abs-1910-07113} and the fine-tuning of large language models \citep{ouyang_training_2022}.
Yet modern RL remains \emph{static}: policies are trained offline with task-specific hyperparameters, so they fail under even minor task variations at deployment \citep{kudithipudi2022biological,DBLP:journals/corr/abs-1911-01547}, or forget past skills when adapted to new tasks \citep{abelDefinitionContinualReinforcement, luoEmpiricalStudyCatastrophic2025}.

\emph{Continual RL} advances a contrasting vision: agents that adapt online in non-stationary environments, without external tuning or human supervision \citep{pan_survey_2025,abelDefinitionContinualReinforcement}.
It requires balancing two competing objectives: retaining prior skills (stability) and acquiring new ones (plasticity) \citep{khetarpal2022towards,pan_survey_2025}.
State-of-the-art policy gradient methods such as Proximal Policy Optimization (PPO) \citep{schulman2017ppo} lose plasticity when the task changes \citep{abelDefinitionContinualReinforcement,aleixoCatastrophicForgettingDeep2023,klein_plasticity_2024}.
A growing body of work identifies various causes of plasticity loss and proposes restoring it by modifying the optimizer \citep{muppidi_fast_2024}, the loss \citep{htang2024mitigating} or the training procedure \citep{sokar_dormant_2023}.
However, recent analyses show that symptoms are hard to tell apart from causes \citep{lyleDisentanglingCausesPlasticity2024}, and that the type of task and of variation strongly affects how methods perform \citep{hirataniDisentanglingMitigatingImpact,wolczykContinualWorldRobotic2021}.
The search for methods with a good stability-plasticity trade-off in RL is thus ongoing.

We here turn to an alternative optimization paradigm: neuroevolution (NE), the evolutionary optimization of neural networks \citep{stanley_designing_2019, risi:book25}.
NE belongs to the family of evolutionary algorithms, black-box optimizers that iteratively refine a population of candidate solutions through mutation and selection \citep{Mitchell1996Introduction,risi:book25}.
We compare two NE algorithms, evolution strategies (ES) \citep{salimans_evolution_2017} and genetic algorithms (GAs) \citep{such_deep_2018}, against PPO \citep{schulman2017ppo}, three state-of-the-art variants for continual RL \citep{htang2024mitigating, muppidi_fast_2024, sokar_dormant_2023} and, to isolate the advantage of a population, population-based RL \citep{jaderberg2017population}.
Across a range of environments and types of change, with policies from a few hundred to more than a million parameters, we find that the GA is the most plastic method and that ES most consistently achieves a good trade-off between stability and plasticity.
PPO, in contrast, loses plasticity, and continual RL variants restore some of it at the expense of stability.


\begin{figure}[t]
    \centering
    \includegraphics[width=\linewidth]{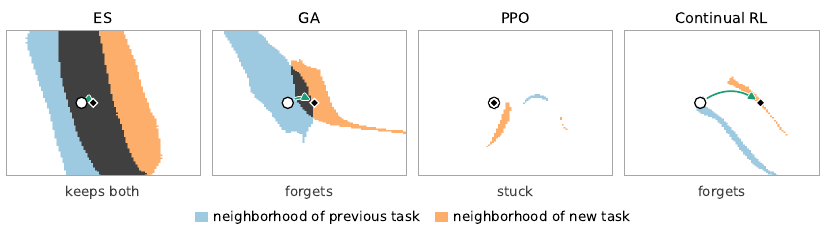}
\caption{\textbf{Measured return landscapes of neuroevolution and reinforcement learning at a task switch.}
Each panel shows the return landscape around a method's solution at a switch between two alternating tasks.
The circle marks the solution of the previous task and the diamond that of the new task.
Blue and orange are the neighborhoods of the previous and the new task, the weights around the solution whose policies still solve it, and dark grey is their overlap. The arrow shows the move from one solution to the other.
 ES has the widest neighborhoods and overlap, and solves both tasks at the switch. The GA's overlap is smaller and at the switch the solution escapes it. PPO has a very small neighborhood and has lost plasticity: it does not move at the switch. Continual RL variants reach the neighborhood of the new task but leave that of the previous one. The panels show empirical landscapes measured in our study (see Section~\ref{sec:basin_measure} for a definition of a neighborhood, Section~\ref{sec:results} for a deeper empirical discussion and Appendix~\ref{app:landscapes} for more examples of such landscapes).}
    \label{fig:overview}
\end{figure}

To understand these differences, we look at the return landscape around the solutions each method finds, the map from policy weights to return \citep{rahnPolicyOptimizationNoisy2023,lehmanESMoreJust2018a}.
We perturb the weights of a solution and call the region in which the perturbed policies still solve the task its \emph{neighborhood}.
\citet{lehmanESMoreJust2018a} showed that ES finds wider neighborhoods than gradient-based optimization, because it maximizes the expected return of perturbed copies of its solution.
Here we further propose that neighborhood size matters for continual learning: when the neighborhoods of two consecutive tasks overlap, a solution to the new task lies close to the current one, so at a task switch a method can reach it while still solving the old task, and be both stable and plastic. We present empirical loss landscapes illustrating this idea in Figure~\ref{fig:overview}. Here the solution of ES and the GA is the centroid of the population and ReDo-PPO \citep{sokar_dormant_2023}, a variant that recycles dormant neurons to counter plasticity loss, stands for a continual RL variant.
We find that ES has the widest neighborhoods and the largest overlap between consecutive tasks, and that the size of this overlap predicts each method's trade-off.

Why does the GA learn new tasks better than ES but forgets more?
We hypothesize that both follow from how the GA selects solutions: where ES averages its perturbed copies into a single centroid, the GA keeps its best individuals and mutates them.
As a result, its population is far more spread out in weight space and its neighborhoods are narrower than those of ES.
This spread is what lets GAs escape local optima \citep{such_deep_2018}: at a task switch, a member far from the current solution may already solve the new task, and selection lets it take over the population.
In a setting with task switches this means that the GA is also more likely to escape a shared neighborhood.
Finally, we show that rewarding behavioural diversity through novelty search \citep{bahlous-boldiDominatedNoveltySearch2025} makes the populatio more plastic, as it becomes more likely that some individual already solves the new task at a switch.

Finally, we show that NE does not exhibit symptoms that accompany plasticity loss in RL: dormant neurons \citep{sokar_dormant_2023} do not accumulate, and weight drift \citep{muppidi_fast_2024} does not cost it plasticity.
This suggests that these symptoms stem from gradient-based optimisation rather than from non-stationarity.

In summary, we make the following observations:
\begin{itemize}[leftmargin=*,noitemsep,topsep=2pt]
    \item ES achieves a good stability-plasticity trade-off most consistently, and the GA is the most plastic method.
    \item ES finds the widest neighborhoods, and the share of a method's neighborhood that solves both consecutive tasks correlates with its trade-off.
    \item Behavioural diversity makes the population more plastic by raising the chance that some member of the population solves the new task.
    \item Dormant neurons do not accumulate in NE, and weight drift does not cost it plasticity.
\end{itemize}
We release a codebase that enables reproducing our experiments: \url{https://github.com/eleninisioti/continual_neuroevolution}.

\section{Background}\label{sec:background}

\subsection{Continual reinforcement learning}\label{sec:problem}
Following previous work \citep{ahn_reset_2024,muppidi_fast_2024}, we define continual RL as a sequence of $T$ RL tasks indexed by $\tau \in \{1, \dots, T\}$. At task $\tau$ the agent interacts with a Markov Decision Process (MDP) $\mathcal{M}_\tau = (\mathcal{S}_\tau, \mathcal{A}_\tau, p_\tau, r_\tau)$ of states, actions, transition probability $p_\tau(s_{t+1} \mid s_t, a_t)$ and reward $r_\tau(s_t, a_t)$. An episode starts from $s_0 \sim \rho_\tau$ and terminates after $H_\tau$ steps or upon a terminal condition, and the agent maximizes the episodic return of each task, $R_\tau = \sum_{t=0}^{H_\tau} r_\tau(s_t, a_t)$. We adopt a boundary-free setting: no information about task change is provided. 

\subsection{Neuroevolution}\label{sec:ne}
Neuroevolution optimises the weights $\theta \in \mathbb{R}^d$ of a policy network $\pi_\theta$ with an evolutionary algorithm \citep{risi:book25,stanley_designing_2019}. Each generation it evaluates $P$ candidate networks, using the episodic return $R_\tau$ as the fitness of each.
A \emph{genetic algorithm} (GA) keeps an archive of $N$ networks. Each generation it perturbs parents drawn uniformly from the archive with Gaussian noise of scale $\sigma$ and keeps the fittest $N$ of archive and offspring \citep{such_deep_2018}.
\emph{Evolution strategies} (ES) keep a single network $\theta$. Each generation they evaluate $P$ perturbations $\theta + \sigma\epsilon_i$, $\epsilon_i \sim \mathcal{N}(0, I)$, and move $\theta$ along their fitness-weighted sum, a Monte Carlo estimate of the gradient of the expected return under $\mathcal{N}(\theta, \sigma^2 I)$ \citep{wierstraNaturalEvolutionStrategies2008,salimans_evolution_2017}.

\section{Related work}

\paragraph{Loss of plasticity in deep RL.}\label{sec:challenges}
Policy-gradient methods produce today's strongest results in domains ranging from robotic locomotion \citep{schulman2017ppo} to LLM fine-tuning \citep{shenfeldRLsRazorWhy2025}.
When the task changes over time, however, they lose plasticity: a network trained on a sequence of tasks learns each new task less well than a freshly initialized one \citep{klein_plasticity_2024,abelDefinitionContinualReinforcement,ahn_reset_2024,julianiStudyPlasticityLoss2024}.
Plasticity plays an important role even within a single task, since RL is inherently non-stationary: the agent's observations and targets depend on its own policy \citep{igl_transient_2020}.
Several causes have been proposed for this loss~\citep{lyleDisentanglingCausesPlasticity2024}.
Neural networks are prone to forgetting earlier knowledge, as updates may overwrite useful weights \citep{mccloskey_catastrophic_1989}.
Gradient-based optimization accumulates \emph{dormant} neurons, units whose activation is near zero on every input and which therefore receive no gradient \citep{sokar_dormant_2023}.
It also lets the weights drift away from their initialization, where the network is most plastic \citep{muppidi_fast_2024}.
An update computed on some inputs also changes the network's outputs on all others, a phenomenon called \emph{churn}, which grows as the representation collapses and gradients on different inputs align \citep{htang2024mitigating}.
Whether these are symptoms or causes of plasticity loss is contested: they co-occur and no single one accounts for the loss \citep{lyleDisentanglingCausesPlasticity2024}.
We benchmark three continual RL variants that address plasticity loss: ReDo, which periodically re-initializes dormant neurons \citep{sokar_dormant_2023}; TRAC, a parameter-free optimizer that adaptively pulls the weights back towards their initialization \citep{muppidi_fast_2024}; and C-CHAIN, which targets the collapse of the representation by reducing churn \citep{htang2024mitigating}.

\paragraph{Population-based RL.}
Population-based training \citep[PBT;][]{jaderberg2017population} trains a population of gradient-based learners in parallel and periodically replaces the weakest members with copies of the strongest. 
We include PBT-PPO to isolate the advantages of a population.
To our knowledge, PBT has not been studied in continual settings.

\paragraph{Evolution in changing environments.}
Evolutionary biology and artificial evolution have long studied how environmental change shapes evolution \citep{grove_evolution_2014,10.1145/3512290.3528826,lehman_extinction_2015}. In these fields change is often seen as a driver rather than a setback. Extinction events can free a population from local optima \citep{lehman_extinction_2015}, and a history of change leaves a population more diverse and better prepared for the next change \citep{grove_evolution_2014,10.1145/3512290.3528826}. 
These studies used either very small networks or much simpler representations.
Quality-Diversity optimization searches for a diverse set of well-performing solutions rather than a single one, measuring diversity in the space of a \emph{behavioural descriptor}, a low-dimensional summary of what a policy does \citep{pughQualityDiversityNew2016}.
Dominated Novelty Search turns novelty into a selection rule, keeping a weak individual if no fitter one behaves like it \citep{bahlous-boldiDominatedNoveltySearch2025}, and AURORA removes the need to hand-design the descriptor by learning it without supervision, as the latent code of an auto-encoder trained on the observations a policy visits \citep{cully_autonomous_2019,grillottiUnsupervisedBehaviorDiscovery2022}.
This diversity makes a robot robust to damage during deployment~\citep{cullyRobotsThatCan2015} and lets NE outcompete RL by escaping local optima~\citep{chalumeau_neuroevolution_2023}.
However, it has not been studied in continual settings, where the archive may need to adapt online. \citet{leuzzi2025} propose an importance-based reward for neuroevolution algorithm, and show how it can be applied in a lifelong settings to control a robotic swarm. \citet{pitzalis2026} apply ES in a lifelong setting for locomotion tasks. Unlike our study, none of them explicitly consider the role of a diverse population of solutions in long continual streams of tasks. \\
Similarly to Quality-Diversity, \citet{nisiotiWhenDoesNeuroevolution2025a} link population diversity to plasticity. They compared NE to RL on transferring skills in a curriculum of tasks and found that RL frequently collapsed at task changes while NE methods reached higher levels of the curriculum. Despite their relevance, these studies considered solely plasticity ignoring forgetting and did not compare NE to continual RL variants. \citet{lillo2026} showed that Quality-Diversity approaches can help foster plasticity and recoverability in MiniGrid environments. \\
Recent studies in LLM fine-tuning remain inconclusive about whether ES-based approaches forget less than RL methods \citep{abdi-etal-2026-evolutionary,qiu2026evolutionstrategiesscalellm}.
Here we measure stability, plasticity and population diversity on a range of RL benchmarks with policies of a few hundred to more than a million parameters. We compare NE to continual RL methods, and propose a mechanism behind NE's advantage.

\paragraph{Neighborhoods in continual loss landscapes}
Studying evolution in switching environments, \citet{wangEvolvingGeneralistsSwitching2019} asked whether a hill climber that alternates between two landscapes can find a local optimum shared by both.
They showed that this is possible through \emph{neighborhood linking}: if the optimum reached on one task lies within a neighborhood of the other, the next switch carries the climber there, and repeated switches chain neighborhoods together until it reaches an optimum shared by both tasks.
We recover this behaviour with neural policies on RL tasks and show that it depends on the width of the neighborhoods and on the selection scheme.
In gradient-based continual learning, \citet{mirzadehUnderstandingRoleTraining2020} linked wider neighborhoods to less forgetting in supervised learning and widened them through the training regime, such as the learning rate and the batch size; here we instead study neighborhood width in RL tasks and focus on how selection under noise increases neighborhood size and affects the stability-plasticity trade-off.

\section{Experimental set-up}

\paragraph{Evaluation metrics}\label{sec:eval_criteria}
All metrics are computed from the return $R_\tau$ (Section~\ref{sec:problem}), averaged over several episodes over multiple environment seeds and deterministic actions.
For the population-based methods (ES, the GA and PBT-PPO) we evaluate the \emph{centroid}, the agent whose parameters are the mean of the population's parameters.
This differs from the usual practice for GAs of reporting the best member: the centroid ensures that every method is compared through a single agent and avoids the upward bias of taking a maximum over many, potentially noisy, evaluations.

For plasticity, we report the \textbf{learning accuracy (LA)}, the return on each task at the end of training on it, and the \textbf{cumulative return (Cum.)} on the current task over training, which also captures how quickly a method recovers after a switch.
For stability, we report \textbf{forgetting (F)}, the return lost on a task between the end of training on it and the end of the run.
We score the stability-plasticity trade-off as LA $-$ F, as is common in continual learning \citep{lopez-paz2017,diaz-rodriguez2018}.
Finally, \textbf{zero-shot transfer (ZT)} is the return on the next task at a switch, before any training on it (see Appendix~\ref{app:metrics} for a formal definition of every metric).


Unless stated otherwise, results report the mean over ten trials with a 95\% percentile-bootstrap confidence interval, and we test differences within a setting with a one-sided Mann-Whitney U test with Holm's correction ($p<0.05$). To put settings on one scale we rescale LA and F per setting so that $0$ is the untrained network and $1$ the best method's mean LA.

\paragraph{Neighborhood width}\label{sec:basin_measure}
The \emph{neighborhood} of a solution is the region of weight space around it in which the policy still solves its task. We say that a policy solves a task when its rescaled return is at least $0.5$. To measure the size of the neighborhood we follow \citet{lehmanESMoreJust2018a}. We take the centroid at the end of a task and perturb the weights of each layer by a random Gaussian direction whose length is a fraction $\epsilon$ of that layer's weight norm, so that the measure does not depend on the scale of the weights \citep{li2018visualizing}. The \emph{neighborhood width} is the radius $\epsilon$ at which half of the perturbed policies no longer solve the task. 

A wide neighborhood ensures that a method solves the new task immediately if it also contains solutions of the new task. The \emph{shared neighborhood} measures this: we perturb the centroid in the same way at a single radius, $\epsilon = 0.1$, and report the fraction of perturbed policies that solve both the current and the next task. 

Appendix~\ref{app:neighborhood} gives the exact procedure, and Appendix~\ref{app:basin_measure} shows that these metrics are robust to other choices of these thresholds.

\paragraph{Population diversity}\label{sec:diversity_measure}
For methods that keep a population of policies we measure how different its members are, both in behavior and weight space. Behavioural diversity is how often two members choose different actions in the same states (for continuous actions, how far apart their actions are). Genomic diversity is the average distance between the weights of two members, normalised by the number of weights so that networks of different sizes can be compared (see Appendix~\ref{app:metrics} for the exact definitions).

\paragraph{Environments and task sequences}\label{sec:tasks}
We consider classic control (CartPole, Acrobot, MountainCar; \citealp{gymnax2022github}), sparse-reward navigation (MiniGrid empty rooms; \citealp{MinigridMiniworld23,nikulin_xland-minigrid_2023}), locomotion (Brax HalfCheetah; \citealp{brax2021github}) and pixel-based physics puzzles requiring million-parameter convolutional networks (Kinetix; \citealp{matthews_kinetix_2024}). To study hard exploration we also consider DeepSea \citep{osband_behaviour_2020}.
To introduce non-stationarity we vary tasks in one of three ways. \emph{Observation offsets} add a fixed random vector to the observation following \citet{muppidi_fast_2024}. \emph{Changes in the transition dynamics} rescale a physical parameter of the agent or environment, change the room size in MiniGrid, or load a new Kinetix level, where both the robot's morphology and the optimal behavior vary widely across tasks. \emph{Action reversal} reverses the order of the discrete actions, or flips the sign of continuous ones.
Every run lasts twenty tasks, in one of two settings, a 20-task setting and  a \emph{two-task} setting, where two tasks alternate ten times. The two settings differ in how much forgetting can be avoided: in a sequence, a single policy can at best stay close to the solutions of neighbouring tasks, so tasks further back are hard not to forget, whereas with two tasks a single policy can show no forgetting and can enable basin linking.
We provide more details about tasks in Appendix~\ref{app:tasks}.

\paragraph{Algorithms}\label{sec:algorithms}
For \textbf{NE} we benchmark ES \citep{salimans_evolution_2017} and a GA \citep{such_deep_2018}.
Unlike ES, which updates its centroid directly, a GA never evaluates its centroid, and when its archive is diverse the mean of its members can score well below the best of them.
For the tasks where this gap appears, we use a simple heuristic: the GA lowers its mutation width while its centroid lags, and a quarter of the offspring keep the original width so that it keeps exploring (see Appendix~\ref{app:ga_focus} for details).
To study the role of population diversity we add Dominated Novelty Search with AURORA descriptors \citep{bahlous-boldiDominatedNoveltySearch2025,grillottiUnsupervisedBehaviorDiscovery2022}, which we call GA + Novelty (see Appendix~\ref{app:aurora} for details).
As \textbf{RL baselines} we benchmark PPO \citep{schulman2017ppo} and three continual variants: TRAC-PPO \citep{muppidi_fast_2024}, ReDo-PPO \citep{sokar_dormant_2023} and C-CHAIN \citep{htang2024mitigating}. To isolate what a population contributes, we also benchmark population-based PPO (PBT-PPO; \citealp{jaderberg2017population}).

Appendix~\ref{app:methods} gives the pseudocode and hyperparameters of every method.
All methods are tuned to solve the stationary version of every environment (see Appendix~\ref{app:noshifts} for their performance without task switches). To check that our conclusions do not depend on one set of hyperparameters, we vary vary hyperparameters for PPO, ES and GA (see Appendix~\ref{app:hyperparameters}), as well as the frequency of task switching (see Appendix~\ref{app:frequency}).

\section{Results}\label{sec:results}
\setcounter{topnumber}{1} 

\begin{figure}[t]
    \centering
    \includegraphics[width=0.8\linewidth]{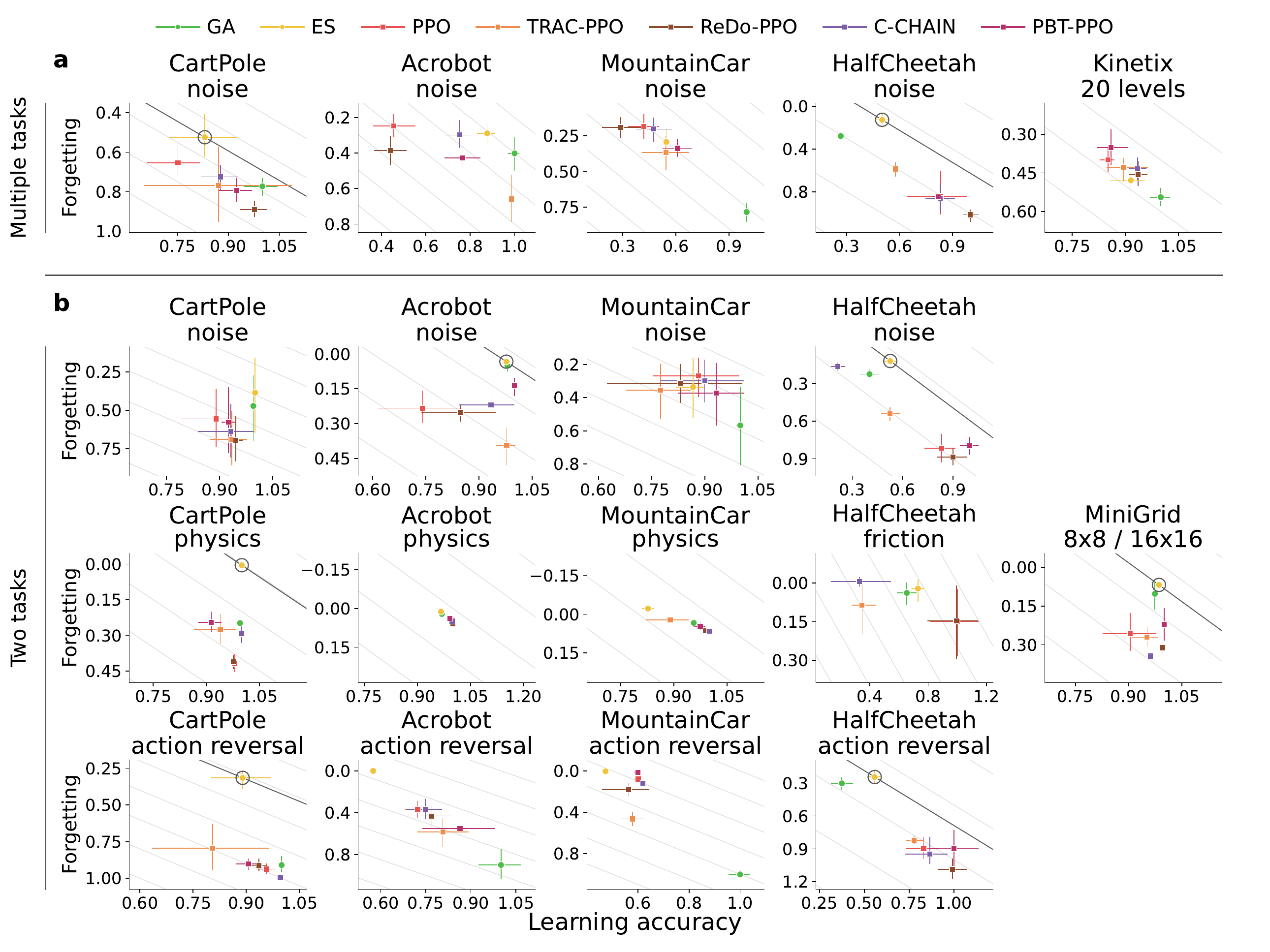}
    \caption{\textbf{Stability-plasticity trade-off.} Learning accuracy (LA, higher is better) against forgetting (F, lower is better) a) with a sequence of twenty tasks and b) with two alternating tasks. Grey lines are equal LA $-$ F. A ring marks the best method on the stability-plasticity trade-off (LA-F). GA is the most plastic method and ES the one with the best trade-off.}
    \label{fig:overall}
\end{figure}

\paragraph{ES achieves the best stability-plasticity trade-off.}
Figure~\ref{fig:overall} shows plasticity (LA) against stability (F) for every method and setting (see Appendix~\ref{app:all_metrics} for corresponding plots of the return over training in every setting).
In each setting, we circle the method with the highest LA $-$ F if it significantly outperforms every method of the other family.
A method that learns only one of two alternating tasks is not eligible.\footnote{Under action reversal: ES on Acrobot and MountainCar, and PPO, C-CHAIN and PBT-PPO on MountainCar.}
ES is circled in eight of the eighteen settings and no RL method in any.
ES falls short in two cases.
Under action reversal on Acrobot and MountainCar, it learns only one ordering of the actions, because at the switch the fitness variance of its population is close to zero (see Appendix~\ref{app:diversity} for the fitness variance over training).
On HalfCheetah, the RL methods converge to higher returns than ES and the GA, also without task switches (see Appendix~\ref{app:noshifts}), so their advantage does not depend on the switches.

\begin{figure}[t]
    \centering
    \includegraphics[width=\linewidth]{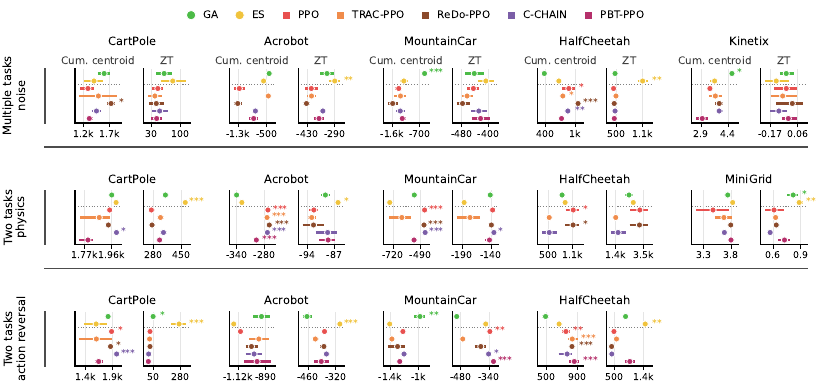}
    \caption{\textbf{Return and zero-shot transfer.} Cumulative return of the centroid (Cum.) and its zero-shot transfer to the next task (ZT), with a sequence of tasks (top) and with two alternating tasks. Mean and 95\% bootstrap interval over trials.}
    \label{fig:metrics}
\end{figure}

\paragraph{The GA is the most plastic method.}
The GA has the highest learning accuracy in ten of the eighteen settings (Figure~\ref{fig:overall}).
Under action reversal on Acrobot and MountainCar it is the only method that fully learns both action orderings: ES freezes on the first ordering it solves in both, and PPO, C-CHAIN and PBT-PPO do so on MountainCar.
The GA forgets more than ES in every setting.
Whether GA or RL forgets more depends on the task: GA forgets more in five settings (MountainCar under noise, over a sequence and with two tasks, Acrobot and MountainCar under action reversal, and Kinetix), where it also learns the most, but less than every RL method in six, among them HalfCheetah under action reversal and under noise over a sequence of tasks, and MiniGrid.

\begin{figure}[t]
    \centering
    \begin{minipage}[b]{0.52\linewidth}
        \centering
        \includegraphics[width=\linewidth]{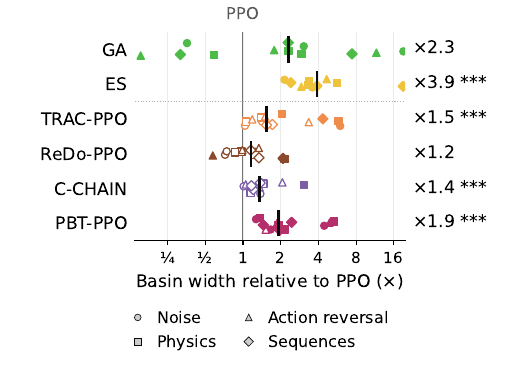}
    \end{minipage}\hfill
    \begin{minipage}[b]{0.44\linewidth}
        \centering
        \includegraphics[width=\linewidth]{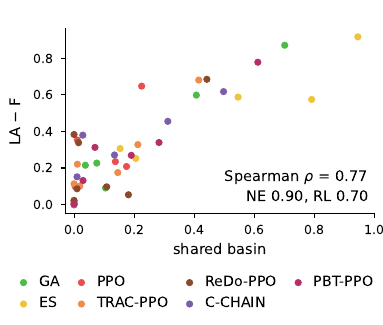}
    \end{minipage}
    \caption{\textbf{Neighborhood width and the shared neighborhood.} Left: neighborhood width relative to PPO, one point per method and classic control or MiniGrid setting; bars are medians and stars a Wilcoxon test over settings. Right: shared neighborhood (random moves of radius $0.1$) against LA $-$ F, with the Spearman correlation over all points and within each family. ES exhibits the largest neighborhood and neighborhood overlap correlates with the stability-plasticity trade-off.}
    \label{fig:basin}
\end{figure}

\paragraph{ES has the highest zero-shot transfer.}
ES has the highest zero-shot transfer in ten of the fourteen settings, significantly above every RL method in eight (Figure~\ref{fig:metrics}).
The GA transfers better than every RL method in five settings (significantly on MiniGrid and CartPole under action reversal), but under action reversal on Acrobot and MountainCar it falls to the return of an untrained policy, having specialised to the ordering it just learned.
The RL methods transfer less, and on HalfCheetah under noise and CartPole under reversal they relearn each task after every switch.
No method transfers to an unseen Kinetix level.

\begin{figure}[t]
    \centering
    \includegraphics[width=0.8\linewidth]{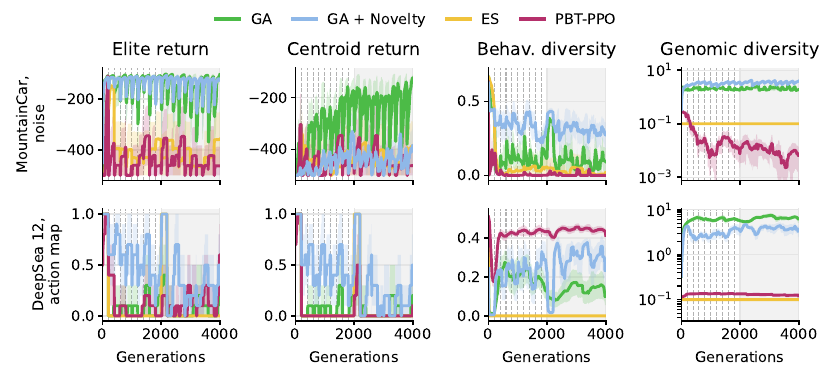}
    \caption{\textbf{Effect of novelty search.} Return of the best individual (elite) and of the centroid, behavioural diversity and genomic diversity; dashed lines mark task switches.}
    \label{fig:novelty}
\end{figure}

\paragraph{Continual RL variants improve plasticity at the expense of stability.}
TRAC-PPO, ReDo-PPO and C-CHAIN keep the network plastic: C-CHAIN learns each task better than PPO in fifteen of the eighteen settings, ReDo-PPO in ten and TRAC-PPO in eight (Figure~\ref{fig:overall}).
Each of the three forgets more than PPO in more than half of the settings (Figure~\ref{fig:overview}; see Appendix~\ref{app:landscapes} for the return landscapes of every method).
The best RL method on the trade-off is PBT-PPO: it has the best LA $-$ F among RL methods in eight settings, against six for C-CHAIN.
When analysing further PBT-PPO (see Appendix \ref{app:pbt}), we see that the main reason for this advantage is reduced per-agent experience: under matched compute its eight members take turns, so each makes an eighth of PPO's updates on a task. This helps where PPO loses plasticity, on Acrobot and MountainCar but hurts on the more challenging HalfCheetah, where an eighth of the updates is too few to learn a task. On the other hand, PBT-PPO does not bring any plasticity benefit: the diversity in its population collapses early (see Figure \ref{fig:novelty}) so it does not exhibit the qualitative behavior that makes GAs plastic.


\paragraph{ES sits in the widest neighborhood, and the shared neighborhood predicts the trade-off.}
ES has the widest neighborhood of all methods, wider than PPO's in every setting, followed by the GA and PBT-PPO (Figure~\ref{fig:basin}, left; see Appendix~\ref{app:basin_test} for the test per environment). Over training the neighborhoods of ES and the GA stay wide, while that of PBT-PPO narrows (see Appendix~\ref{app:basin_evolution} for the width over training).
Width is measured in random directions, the moves NE makes, but RL follows the gradient of the return, so its neighborhood could be wider along its own path \citep{rahnPolicyOptimizationNoisy2023}. To account for this, we repeat the neighborhood analysis along the updates each RL method actually makes, and show that the neighborhood of PPO is as narrow along its path as in random directions. Only the neighborhood of ReDo-PPO is wider along its path, about twice as wide as in random directions (see Appendix~\ref{app:shared_basin} for this analysis).
The shared neighborhood (Section~\ref{sec:basin_measure}) predicts LA $-$ F, across methods and within each family.

\begin{figure}[!t]
    \centering
    \begin{minipage}[c]{0.655\linewidth}
        \centering
        \includegraphics[width=\linewidth]{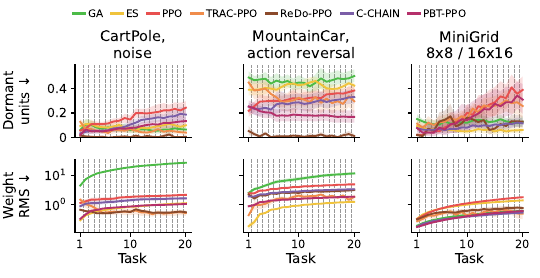}
    \end{minipage}\hfill
    \begin{minipage}[c]{0.325\linewidth}
        \caption{\textbf{Dormant units and weight drift.} Fraction of dormant hidden units of the centroid (top) and root mean square of its weights (bottom). Dashed lines mark task switches. The dormant fraction of NE depends on the task and stays constant; that of RL rises.}
        \label{fig:plasticity_symptoms}
    \end{minipage}
\end{figure}

\paragraph{Novelty search makes populations more plastic.}
The GA keeps a diverse population, which can help it escape a local optimum after a switch or raise the chance that some individual already lies in the neighborhood of the new task. We test this link between diversity and plasticity by rewarding diversity explicitly through novelty search \citep{bahlous-boldiDominatedNoveltySearch2025}. We choose two settings where a switch can trap the whole population in a local optimum (described in Appendix~\ref{app:tasks}): MountainCar, whose sparse reward gives no signal until the car reaches the flag, under stronger observation noise, and DeepSea, where each task changes the action map of the single rewarded path.
GA + Novelty keeps a more behaviourally diverse population in both, and its best individual gains cumulative return in each (Figure~\ref{fig:novelty}); on DeepSea it re-finds the new action map after $57\%$ of the switches, against $11\%$ for the GA.
The centroid does not track this gain: it averages over members that solve different tasks, and the average of their weights solves none of them.
Diversity thus buys plasticity for a method that can pick its best individual, moving the GA towards the plastic end of the trade-off (see Appendix~\ref{app:novelty} for more task settings).

\paragraph{NE does not show the symptoms of plasticity loss of RL.}
PPO's dormant fraction keeps rising over task switches, and ReDo-PPO, which recycles dormant units, keeps it the lowest (Figure~\ref{fig:plasticity_symptoms}, top).
The dormant fraction of NE depends on the task but does not increase at task switches (see Appendix~\ref{app:plasticity_all} for all tasks).
As NE methods do not learn through gradients, it is not surprising that they perform well in the presence of dormant neurons.
What our experiments suggests, however, is that non-stationarity alone is not sufficient to lead to their increase and that this increase may appear only under gradient-based optimization.
We also observe that 
weight drift \citep{muppidi_fast_2024} does not correlate to plasticity loss in NE either (Figure~\ref{fig:plasticity_symptoms}, bottom). Fo ES and the continual RL variants weights remain smaller for PPO, but for the GA, the most plastic method, weights can drift by orders of magnitude.

\setcounter{topnumber}{2}

\section{Discussion}\label{sec:discussion}
We compared two neuroevolution methods, ES and a GA, to PPO and its continual variants across environments and kinds of task change, with matched compute and no signal of a switch. ES achieves the best stability-plasticity trade-off.
The GA is the most plastic method in almost every environment, and rewarding novelty increases its plasticity further.
PPO loses plasticity in most environments, and its continual variants restore it at the cost of stability.
Neuroevolution does not show the symptoms of plasticity loss reported in continual RL: its networks do not accumulate dormant units, and the GA's weights drift further than PPO's without costing it plasticity, which suggests that these symptoms stem from gradient-based optimisation rather than from the non-stationarity.

We propose a mechanism explaining these results: exploring in parameter space biases learning towards wide neighborhoods, which are more likely to overlap with the neighborhood of the next task: across methods, the size of this shared neighborhood predicts the stability-plasticity trade-off. This builds on evidence that wider neighborhoods reduce forgetting\citep{mirzadehUnderstandingRoleTraining2020} and that ES prefers robust solutions \citep{lehmanESMoreJust2018a}. To the best of our knowledge, we are the first to propose parameter-space exploration as a mechanism for balancing stability and plasticity in continual RL. 

Despite these benefits, NE is also characterized by limitations that may lead to it underperforming RL in some settings. First, it is less sample-efficient, and complex tasks often require large populations \citep{such_deep_2018}. 
Second, where good solutions are finely tuned, as in HalfCheetah in our experiments, NE does not match RL's stationary performance: gradient steps can climb narrow peaks of the return, which ES smooths over \citep{lehmanESMoreJust2018a}.
Yet ES has recently been shown impressing scaling, fine-tuning language models with billions of parameters \citep{qiuEvolutionStrategiesScale2025,sarkarEvolutionStrategiesHyperscale2025}.
Our results are also encouraging: both ES and the GA learn the Kinetix levels with a convolutional policy of more than a million parameters better than RL at the same budget.



Our results open two directions for future work.
First, the two NE mechanisms we found beneficial for continual learning, parameter-space exploration for stability and population diversity for plasticity, may also benefit RL.
For example, injecting noise into the policy's weights \citep{plappert2018parameter} could let PPO follow the gradient of a smoothed return, and population-based methods that reward diversity \citep{NEURIPS2020_d1dc3a82} could keep it plastic.
Second, NE itself can be improved with mechanisms we left out.
For example, ES and GAs can adapt the range of their mutations online \citep{back1992self,hansen2016cma,wierstraNaturalEvolutionStrategies2008}; raising the mutation rate when performance drops has long been used to track changing environments \citep{Grefenstette1992GeneticAF,pedersenHypernetworksThatEvolve2025}.
Overall, our work introduces a long-standing optimization paradigm to continual RL and proposes novel mechanisms for balancing stability and plasticity.

\section*{Acknowledgements}
This project was supported by a European Research Council (ERC) grant (GA no.~101045094, project ``GROW-AI'').

\mainreferences

\beginappendixpart
\appendix
\renewcommand{\topfraction}{0.95}
\renewcommand{\bottomfraction}{0.9}
\renewcommand{\textfraction}{0.05}
\renewcommand{\floatpagefraction}{0.8}
\setcounter{topnumber}{3}
\setcounter{bottomnumber}{2}
\setcounter{totalnumber}{4}
\makeatletter\setlength{\@fptop}{0pt}\makeatother   
\addtocontents{toc}{\protect\setcounter{tocdepth}{3}}
\renewcommand{\contentsname}{Appendix}
\tableofcontents
\clearpage
\raggedbottom   
\section{Implementation details}
\label{app:implementation}

\subsection{Environment suites and their continual variants}
\label{app:tasks}

We evaluate on four environment suites, chosen to span a wide range of learning
challenges: gymnax \citep{gymnax2022github} for low-dimensional
control, MiniGrid \citep{MinigridMiniworld23} for sparse-reward
navigation, Brax \citep{brax2021github} for high-dimensional robot locomotion, and Kinetix~\citep{matthews_kinetix_2024} for pixel-based
control of procedurally generated rigid-body scenes.

Table~\ref{tab:tasks} gives the observation space, the action space, the number of policy parameters and the episode length of every environment.

\begin{table}[!htbp]
\centering
\setlength{\belowcaptionskip}{6pt}
\caption{The environments. Observations are real-valued ($\mathbb{R}^{n}$) or
binary ($\{0,1\}^{n}$): DeepSea is a one-hot position on the grid, MiniGrid a
one-hot encoding of the tile and colour in each cell of the egocentric view,
and Kinetix an RGB image.  $\mathcal{D}(k)$: $k$ discrete actions. Policy params
count the actor only. Steps / task is the environment-step budget of one task.}
 \label{tab:tasks}
\small
\setlength{\tabcolsep}{4pt}
\begin{tabular}{llrrrrr}
\toprule
Suite & Environment & Obser- & Actions & Policy & Episode & Steps / \\
      &      & vations & & params & length & task \\
\midrule
gymnax   & CartPole-v1     & $\mathbb{R}^{4}$     & $\mathcal{D}(2)$  & $386$       & $500$  & $1.54{\times}10^{8}$ \\
gymnax   & Acrobot-v1      & $\mathbb{R}^{6}$     & $\mathcal{D}(3)$  & $435$       & $500$  & $1.54{\times}10^{8}$ \\
gymnax   & MountainCar-v0  & $\mathbb{R}^{2}$     & $\mathcal{D}(3)$  & $371$       & $500$  & $1.54{\times}10^{8}$ \\
gymnax   & DeepSea ($12{\times}12$) & $\{0,1\}^{144}$ & $\mathcal{D}(2)$  & $2{,}626$   & $12$   & $3.69{\times}10^{6}$ \\
MiniGrid & Empty rooms     & $\{0,1\}^{7\times7\times25}$ & $\mathcal{D}(6)$ & $7{,}758$  & $1024$ & $3.15{\times}10^{8}$ \\
Brax     & HalfCheetah  & $\mathbb{R}^{17}$    & $[-1,1]^{6}$      & $19{,}590$  & $1000$ & $2.46{\times}10^{7}$ \\
Kinetix  & \texttt{m/h0}--\texttt{h19} & $\mathbb{R}^{125\times125\times3}$ & $\mathcal{D}(3)^{4}{\times}\mathcal{D}(2)^{2}$ & $1{,}128{,}256$ & $128$ & $1.31{\times}10^{7}$ \\
\bottomrule
\end{tabular}
\end{table}

Every task in a sequence lasts the same number of environment steps; at the end of
each task the environment changes.
No method is told when a switch occurs.
For NE a task costs
$G \times P \times R \times T$ environment steps, where $G$ is the number of
generations per task, $P$ the population size, $R$ the number of episodes per
candidate and $T$ the episode length; for RL methods it costs $U \times N \times L$ steps, with $U$ updates per task, $N$ parallel environments and $L$ the rollout length.
Table~\ref{tab:architectures} gives the networks.

\begin{table}[!htbp]
\centering
\setlength{\belowcaptionskip}{6pt}
\caption{Policy and value network architectures.  Hidden widths are listed in
order and the activation column is the \emph{hidden} activation; every layer
carries a bias.  The policy is shared by all methods; the value network exists
only for the gradient-based methods.
On HalfCheetah the PPO
actor is the evolved policy plus a single state-independent $\log\sigma$
vector (initialised at $-0.5$), adding $6$ parameters; on
the discrete-action environments the PPO actor \emph{is} the evolved policy, with a
categorical head.}
\label{tab:architectures}
\small
\setlength{\tabcolsep}{4pt}
\begin{tabular}{llllrlr}
\toprule
 & \multicolumn{4}{c}{Policy network (all methods)} & \multicolumn{2}{c}{Value network (RL only)} \\
\cmidrule(lr){2-5}\cmidrule(lr){6-7}
Environment & hidden & act. & output & params & hidden & params \\
\midrule
CartPole-v1    & $16, 16$       & ReLU & linear    & $386$           & $128,128,128$ & $33{,}793$ \\
Acrobot-v1     & $16, 16$       & ReLU & linear    & $435$           & $128,128,128$ & $34{,}049$ \\
MountainCar-v0 & $16, 16$       & ReLU & linear    & $371$           & $128,128,128$ & $33{,}537$ \\
MiniGrid       & conv $4$; $64$ & ReLU & linear    & $7{,}758$       & $128,128,128$ & $190{,}081$ \\
HalfCheetah    & $128, 128$     & tanh & $\tanh$   & $19{,}590$      & $128,128$     & $18{,}945$ \\
Kinetix        & conv $16,32$; $128^{\times 5}$ & tanh & linear & $1{,}128{,}256$ & $128^{\times 5}$ & $1{,}115{,}009$ \\
\bottomrule
\end{tabular}
\end{table}

\subsubsection{Classic control and DeepSea (gymnax)}

\paragraph{Task set-up}
We use three classic control environments from gymnax \citep{gymnax2022github}:
\textit{CartPole-v1}, balancing a pole on a moving cart; \textit{Acrobot-v1},
swinging a two-link pendulum above a height; and \textit{MountainCar-v0},
driving an under-powered car up a hill.  The episode length is $500$ steps.
MountainCar has a sparse reward: it pays a constant penalty per step until the
car reaches the flag, so a policy that never reaches it receives no learning
signal, as in MiniGrid below.
For the novelty comparison (Figure~\ref{fig:novelty}) we add
\textit{DeepSea} \citep{osband_behaviour_2020}, an exploration environment in which
the agent descends a $12 \times 12$ grid and is rewarded only for reaching the
bottom-right cell; we drop bsuite's small penalty per right move, so every
other trajectory scores $0$. It is the hardest exploration problem we
consider: the reward needs the right action at all twelve steps, so a random
policy finds it once in $2^{12} = 4096$ episodes.

\paragraph{Network architecture}
The policy is a multi-layer perceptron with two hidden layers of $16$ ReLU
units and a linear output layer producing the action logits.

\paragraph{Continual variants}
Every task sequence is $20$ tasks of $200$ generations each, and the first
task is always the unmodified environment.
\begin{itemize}
\item \emph{Observation noise} \citep{muppidi_fast_2024}: a task adds a
fixed offset, drawn from $\mathcal{N}(0, \sigma^{2} I)$, to the observation.
With ten tasks, each visited twice, $\sigma = 1$ on CartPole and Acrobot
and $\sigma = 0.1$ on MountainCar, where $\sigma = 1$ is too high relative to
the range of the observations; \citet{htang2024mitigating} reduce the noise on
MountainCar for the same reason, to $\sigma = 0.02$.
With two alternating tasks the offset is halved, $\sigma = 0.5$ and $0.05$: this family asks whether a method holds one policy that solves
both tasks, so the offset is kept small enough for such a policy to exist.
\item \emph{Physics}: a task rescales the pole length (CartPole), the link
masses (Acrobot) or gravity (MountainCar).  With ten tasks the multiplier
is drawn log-uniformly in $[0.5, 2]$, $[1/1.15, 1.15]$ and $[1/1.5, 1.5]$;
with two, it is fixed at $\times 3$, $\times 1.15$ and $\times 1.5$.
\item \emph{Action reversal}: two alternating tasks, the second with the
order of the discrete actions reversed.  On DeepSea a task is instead a
fresh random map of which action moves right in each cell; ten maps are each
visited twice.
\end{itemize}

\subsubsection{MiniGrid rooms}

\paragraph{Task set-up}
We use the JAX reimplementation of MiniGrid \citep{MinigridMiniworld23} in
xminigrid \citep{nikulin_xland-minigrid_2023}, in which the agent must reach a
goal in a room from a $7\times7$ egocentric view, with a sparse reward that is
higher for shorter paths.  We use the empty rooms \texttt{EmptyRandom-8x8}
and \texttt{EmptyRandom-16x16}, with their step caps of $256$ and $1024$.

\paragraph{Network architecture}
The policy is the small convolutional network of \citet{young19minatar}: one
$3\times3$ convolution with $4$ channels, a dense layer of $64$ units, ReLU
activations and a linear output layer producing the six action logits.

\paragraph{Continual variants}
A task is the room being played: the $8\times8$ and $16\times16$ rooms
alternate over $20$ tasks of $200$ generations each.

\subsubsection{Robot locomotion (Brax HalfCheetah)}

\paragraph{Task set-up}
We use \textit{HalfCheetah} in Brax \citep{brax2021github} with the MJX
backend, a planar biped that must run forward.  The reward is a Gaussian of
the forward speed centred on a target speed of $10$, so the task asks for a
specific gait rather than maximum velocity.  The episode length is $1000$
steps.

\paragraph{Network architecture}
The policy is a multi-layer perceptron with two hidden layers of $128$
$\tanh$ units and a $\tanh$ output layer that squashes the actions into the
actuator range.

\paragraph{Continual variants}
Every task sequence is $20$ tasks long, and the first task is the
unmodified body.
\begin{itemize}
\item \emph{Observation noise}: a fixed observation offset drawn from
$\mathcal{N}(0, 0.5^{2} I)$, either ten tasks each visited twice or two
alternating tasks.
\item \emph{Physics}: two alternating tasks, the second rescaling the
ground friction by a multiplier drawn per trial log-uniformly in $[0.2, 5]$.
PBT-PPO is not run in this setting.
\item \emph{Action reversal}: two alternating tasks, the second flipping
the sign of the action.
\end{itemize}

\subsubsection{Kinetix: twenty rigid-body levels}

\paragraph{Task set-up}
Kinetix \citep{matthews_kinetix_2024} is a JAX rigid-body engine in which a
level is a scene of shapes, joints, motors and thrusters, and the goal is to
bring two objects into contact.  We use the twenty hand-designed
medium-difficulty levels: unicycle, car left, car right, car thrust,
thrust the needle, angry birds, thrust over, car flip, weird vehicle, spin
the right way, thrust right easy, thrust left easy, thrustfall left,
thrustfall right, thrustblock, thrustshoot, thrustcontrol right,
thrustcontrol left, thrust right very easy and thrust left very easy. They
are listed in the benchmark's numbering, \texttt{h0} to \texttt{h19}, which
is also the order in which a task sequence visits them. The numbering does not
rank the levels by difficulty: all twenty belong to the benchmark's medium
set, the two easiest come last, and the stationary returns show no trend with
the index (Figure~\ref{fig:noshifts_all}).
We employ pixel observations,
multi-discrete actions (four ternary motors and two binary thrusters), a frame
skip of $2$, a dense-reward scale of $0.2$ and an episode length of $128$
steps.

\paragraph{Network architecture}
We use the \texttt{model-base} network of \citet{matthews_kinetix_2024}, run
non-recurrently: two convolutions ($16$ channels, $8\times8$, stride $4$; $32$
channels, $4\times4$, stride $2$), a five-layer $\tanh$ MLP of width $128$ and
one categorical head per motor or thruster.  PPO uses their full actor--critic
network, which adds a value MLP of the same shape on the shared convolutional
embedding.

\paragraph{Continual variants}
A task is the level being played: the twenty levels are presented once
each, in the order listed above, for $200$ generations each.

\subsection{Methods}
\label{app:methods}

All methods train the parameters $\theta \in \mathbb{R}^{d}$ of the same policy
network $\pi_{\theta}$. $F(\theta)$ is the expected return of $\pi_{\theta}$ on the current
task.

\subsubsection{Neuroevolution: ES, GA and GA + Novelty}

Every NE method is an instance of the ask-tell loop of
Algorithm~\ref{alg:ne} \citep{evosax2022github}, and is defined by
its $\textsc{Ask}$ and $\textsc{Tell}$. A candidate's fitness is its mean
return over $R$ episodes, so every method spends $PR$ episodes per generation.

\begin{algorithm}[t]
\caption{The neuroevolution loop. $Z$ holds any per-candidate quantity that $\textsc{Tell}$ needs}
\label{alg:ne}
\begin{algorithmic}[1]
\Require initial search state $S$, population size $P$, episodes per
candidate $R$
\For{$g = 1, 2, \dots$}
  \State $(X, Z) \gets \textsc{Ask}(S)$
         \Comment{$X = (x_{1}, \dots, x_{P})$, $x_{i} \in \mathbb{R}^{d}$}
  \State $f_{i} \gets \dfrac{1}{R}\sum_{r=1}^{R}
         \mathrm{Return}\big(\pi_{x_{i}}\big)$ on the current task,
         $i = 1, \dots, P$
  \State $S \gets \textsc{Tell}(S, X, Z, f)$
\EndFor
\end{algorithmic}
\end{algorithm}

\paragraph{ES}
Algorithm~\ref{alg:es} gives the ES of Section~\ref{sec:ne}. Perturbations come in antithetic pairs, and the shaped
fitness is the z-score $u = (f - \bar{f})/\mathrm{std}(f)$, stepped with SGD
\citep{salimans_evolution_2017}. On Kinetix we instead use centred ranks,
$u_{i} = \mathrm{rank}(f_{i})/(P-1) - 1/2$, and Adam, as in evosax's
\texttt{Open\_ES}. $\sigma$ is not adapted.

\begin{algorithm}[t]
\caption{$\textsc{Ask}$ and $\textsc{Tell}$ for ES.}
\label{alg:es}
\begin{algorithmic}[1]
\Require state $S = (\theta, \text{optimiser state})$, width $\sigma$,
learning rate $\alpha$, population size $P$ (even)
\Procedure{Ask}{$S$}
  \State $\epsilon_{i} \sim \mathcal{N}(0, I_{d})$, $i = 1, \dots, P/2$;
         \quad $\epsilon_{P/2+i} \gets -\epsilon_{i}$
         \Comment{antithetic pairs}
  \State \Return $X = (\theta + \sigma\epsilon_{i})_{i=1}^{P}$,
         $Z = (\epsilon_{i})_{i=1}^{P}$
\EndProcedure
\Procedure{Tell}{$S, X, \epsilon, f$}
  \State $u \gets \mathrm{shape}(f)$
         \Comment{z-score; centred ranks on Kinetix}
  \State $g \gets \dfrac{1}{P\sigma}\sum_{i=1}^{P} u_{i}\,\epsilon_{i}$
         \Comment{gradient of the smoothed objective}
  \State $\theta \gets \theta + \textsc{OptStep}(\alpha, g)$
         \Comment{SGD; Adam on Kinetix}
\EndProcedure
\end{algorithmic}
\end{algorithm}

\paragraph{GA}
Algorithm~\ref{alg:ga} gives the GA of Section~\ref{sec:ne}
\citep{such_deep_2018}, with an archive $A$ of $\mu$ genomes and $P - \mu$
offspring per generation, no crossover and $(\mu + \lambda)$ truncation
selection. The archive is re-scored
every generation, so every fitness in a selection comes from the same
task. The archive counts towards $P$, so a generation still costs $PR$
episodes.

\paragraph{GA variant for tracking elite}\label{app:ga_focus}
The agent we report for the GA is its centroid, the mean of the archive
(Section~\ref{sec:eval_criteria}). In most environments the archive is a tight cluster
and its mean scores as well as its best member. On Kinetix and MountainCar it
is not: the archive is spread over several solutions, and their mean solves
nothing. On stationary MountainCar the centroid of the plain GA stays at the
return floor of $-500$ in half the seeds while its best member reaches the
goal. We think the reason is that many policies tie on these two
environments: on MountainCar every policy that never reaches the goal scores
the floor, and on Kinetix every policy that solves a level scores close to the
same. Selection cannot choose between tied members, and a child that only
equals its parent replaces it, so the archive drifts apart across the region
of equal fitness. The mean of members spread over that region need not lie
inside it.
On these two environments we therefore add a heuristic rule that pulls
the archive together whenever its centroid falls behind. One of the $P$
evaluations of each generation scores the centroid $\bar{\theta}_{t}$, and
after selection the rule measures the share of archive members that the
centroid scores at least as well as,
\[
  p_{t} = \frac{1}{\mu}\,
          \big|\{\, i : f(\bar{\theta}_{t}) \geq f(a_{i}) \,\}\big|.
\]
Two quantities follow $p_{t}$ towards a target $p^{\ast} = 0.9$: the mutation
width $\sigma_{t}$, and the focus $\omega_{t}$, the share of the archive, best
first, that offspring are bred from,
\[
  \sigma_{t+1} = \operatorname{clip}\big(\sigma_{t}\,
                 e^{\eta_{\sigma}(p_{t} - p^{\ast})},\;
                 \sigma_{\min},\; \sigma_{0}\big),
  \qquad
  \omega_{t+1} = \operatorname{clip}\big(\omega_{t}\,
               e^{\eta_{\omega}(p_{t} - p^{\ast})},\;
               1/\mu,\; 1\big),
\]
with $\omega_{0} = 1$, $\eta_{\sigma} = 0.1$, $\eta_{\omega} = 0.3$ and
$\sigma_{\min} = 10^{-5}$. Each offspring's parent is drawn uniformly from the
best $\lceil \omega_{t}\mu \rceil$ members of the archive. While the centroid
lags ($p_{t} < p^{\ast}$) both shrink, faster the further it lags, towards a
single parent and the floor $\sigma_{\min}$; once it catches up they grow back
to the whole archive and the initial width $\sigma_{0}$. These settings were
chosen by hand in short pilots, not tuned. In addition, a quarter
of the offspring are always bred from the whole archive at $\sigma_{0}$. This
ensures that the GA can still search after a task switch: once the archive has
consolidated on one solution, all its members fail the new task alike and
$\sigma$ would keep shrinking. An explorer that solves the new task
outranks them and becomes the parent of the next consolidation. While the
centroid tracks, explorers fail and never displace a solver. The rule reads
only the fitnesses of the current generation, never the task index or the
time of a switch.

\paragraph{GA + Novelty}
GA + Novelty (Section~\ref{sec:ne}) ranks the archive by dominated novelty
$d_{i}$ instead of fitness (Algorithm~\ref{alg:ga}). With $b_{i}$ the
behavioural descriptor of individual $i$ and $\mathcal{N}_{k}(i)$ its $k$
nearest neighbours in descriptor space among the individuals at least as fit,
\[
  d_{i} = \frac{1}{k}\sum_{j \in \mathcal{N}_{k}(i)}
          \lVert b_{i} - b_{j} \rVert,
\]
and $d_{i} = +\infty$ when no individual is at least as fit. We differ from
\citet{bahlous-boldiDominatedNoveltySearch2025} in one detail: we rank every
individual with no strictly fitter one first, whereas the reference protects
only the single fittest. When several individuals tie at the top with the same
descriptor, as all DeepSea solvers do, their $d_{i}$ is $0$ and the reference
would drop them all. We use $k = 3$ as in the reference and $\mu = P/2$.

\paragraph{Learned descriptors}\label{app:aurora}
As in the unsupervised variant of the reference, the descriptor $b_{i}$ is
learned with AURORA
\citep{cully_autonomous_2019,grillottiUnsupervisedBehaviorDiscovery2022}: it
is the $6$-dimensional latent code of an LSTM auto-encoder trained on the
observations individual $i$ visits in one episode (sub-sampled to $50$ steps,
before any observation offset). The auto-encoder is refit on the archive's
trajectories at generations $0$ and $8\,n(n+1)/2$, $n = 1, 2, \dots$ ($31$
refits in $4000$ generations), and the archive is re-encoded after each refit.
The schedule depends only on the generation count, so GA + Novelty receives no
information about switches (Section~\ref{sec:algorithms}). Each fit runs Adam
(learning rate $10^{-3}$, batch $128$) for $100$ epochs the first time and
$25$ afterwards.

\begin{algorithm}[t]
\caption{$\textsc{Ask}$ and $\textsc{Tell}$ for the GA. GA + Novelty changes
only the ranking in $\textsc{Tell}$.}
\label{alg:ga}
\begin{algorithmic}[1]
\Require archive $A$ of $\mu$ genomes, population size $P$, mutation scale
$\sigma$
\Procedure{Ask}{$A$}
  \State $x_{j} \gets a_{j} + \sigma\,\epsilon_{j}$, \;
         $a_{j} \sim \mathrm{Uniform}(A)$, \;
         $\epsilon_{j} \sim \mathcal{N}(0, I_{d})$, \; $j = 1, \dots, P - \mu$
  \State \Return $X = (x_{1}, \dots, x_{P-\mu}) \cup A$, \; $Z = \emptyset$
         \Comment{the archive is re-scored}
\EndProcedure
\Procedure{Tell}{$A, X, \cdot, f$}
  \State $A \gets$ the $\mu$ members of $X$ ranked highest by $f_{i}$
  \Statex \hspace{\algorithmicindent}\textbf{GA + Novelty:}
         rank by $d_{i}$ instead, with the top fitness tier first
\EndProcedure
\end{algorithmic}
\end{algorithm}

\subsubsection{Gradient-based methods}

\paragraph{PPO}\citep{schulman2017ppo} is a policy-gradient method: it
makes actions more likely in proportion to their \emph{advantage}
$\hat{A}_{t}$, an estimate of how much better taking action $a_{t}$ in state
$s_{t}$ turned out than the policy's average behaviour in that state, i.e.\ the
return that followed minus the value $V_{\psi}(s_{t})$ the critic predicted
there.  To keep each update close to the policy that collected the data, PPO
clips the ratio $\pi_{\phi}(a_{t} \mid s_{t}) /
\pi_{\phi_{\mathrm{old}}}(a_{t} \mid s_{t})$ to $[1-\varepsilon,
1+\varepsilon]$ in its objective $\mathcal{L}^{\mathrm{clip}}$, which lets it
reuse each rollout for several epochs of minibatch updates
(Algorithm~\ref{alg:ppo}).

On the continuous-action environments the actor adds a
state-independent $\log\sigma$ vector and observations are normalised by a
running estimate; on the discrete-action environments the head is categorical and
observations are used raw.
Algorithm~\ref{alg:ppo} gives the PPO update on which all continual variants build.

\begin{algorithm}[t]
\caption{PPO.  One update: $K$ epochs of minibatch ascent on a single rollout.}
\label{alg:ppo}
\begin{algorithmic}[1]
\Require policy $\pi_{\phi}$, value $V_{\psi}$, $N$ environments, rollout
length $L$, epochs $K$, minibatches $M$
\State collect $N \times L$ transitions with $\pi_{\phi}$, continuing the
       environments from the state the previous update left them in
\State $\hat{A} \gets \mathrm{GAE}(\gamma, \lambda)$, bootstrapped with
       $V_{\psi}(s_{L})$
\For{$K$ epochs, over $M$ minibatches each}
  \State $\mathcal{L} \gets \mathcal{L}^{\mathrm{clip}}(\phi)
         + c_{v}\mathcal{L}^{V}(\psi) - c_{H}\mathcal{H}[\pi_{\phi}]$
  \State $(\phi, \psi) \gets \mathrm{Adam}(\alpha)$ step on $\mathcal{L}$,
         with gradients clipped to $\lVert g \rVert \leq g_{\max}$
\EndFor
\end{algorithmic}
\end{algorithm}

\paragraph{TRAC} \citep{muppidi_fast_2024} wraps the Adam step in line~5 of Algorithm~\ref{alg:ppo}; we use the authors' implementation.
Write $\theta$ for the parameters of both networks, $\theta_{0}$ for their
value at initialisation, and $\tilde{\theta}_{t}$ for Adam's own iterate.
Adam runs as usual on $\tilde{\theta}_{t}$, but the gradients it receives are
computed at the TRAC iterate
\[
  \theta_{t} = \theta_{0} + S_{t}\,(\tilde{\theta}_{t} - \theta_{0}),
\]
which keeps an $S_{t}$ fraction of the displacement Adam has accumulated since
initialisation.  A parameter-free tuner sets $S_{t}$ online from the inner
product $\langle g_{t}, \theta_{t} - \theta_{0} \rangle$ between each gradient
and the current displacement; it sums six copies with discount factors
$0.9$ to $0.999999$ and starts at $S \approx 0$.  $S_{t}$ typically stays in
$[0, 1]$, so $\theta_{t}$ is a convex combination of Adam's iterate and the
initialisation, and $S_{t} \equiv 1$ recovers Adam.  Unlike an $L_{2}$
penalty towards $\theta_{0}$, the pull has no coefficient to tune and its
strength follows the data.

\paragraph{ReDo}\label{app:redo} \citep{sokar_dormant_2023} adds a periodic pass after line~6.  Every
$T_{\mathrm{redo}}$ updates, each hidden neuron is scored on a probe batch, the
score is normalised by its layer's mean, and every neuron scoring at or below
$\tau$ is reset: its incoming weights are resampled from the layer's
initialiser, its bias and its outgoing weights are set to zero, and the Adam
moments and bias-correction count of every weight touched are cleared.  Because
the outgoing weights are zeroed, the network's function is unchanged at the
moment of the reset.  The score is the neuron's mean absolute activation in a
ReLU network; under $\tanh$ we use the standard deviation of the activation
instead, since a $\tanh$ unit saturated at $\pm 1$ transmits no gradient and
yet attains the largest possible mean absolute activation.  Both scores are
normalised by their layer's mean, so a single $\tau$ applies to either.

\paragraph{C-CHAIN} \citep{htang2024mitigating} adds one term to line~4.  The policy is
regularised towards the policy of one gradient step earlier, evaluated on a
minibatch of the same rollout drawn independently of the training minibatch,
\[
  \mathcal{L} \gets \mathcal{L}
  + \lambda_{t}\,
  \mathbb{E}_{s \sim \mathcal{B}^{\mathrm{reg}}}
  \big[ H\big(\pi_{\phi_{t-1}}(\cdot \mid s), \pi_{\phi}(\cdot \mid s)\big)
  \big],
\]
where $\lambda_{t} \geq 0$ is a coefficient that weights the regulariser
against the PPO loss, and $H$ is the cross-entropy on the discrete-action environments and
the squared difference of the action means on the continuous ones.  The
coefficient is not constant: it is set so that the weighted regulariser is a
fixed fraction $\kappa$ of the policy loss,
$\lambda_{t} = \max\big(\kappa\,\overline{|\mathcal{L}^{\pi}|} /
\overline{\mathcal{L}^{\mathrm{reg}}}, \; \lambda_{\min}\big)$, with both
losses averaged over a window of past updates, and it is applied only after a
warm-up.  The term suppresses the off-diagonal entries of the empirical NTK
(Eq.~\ref{eq:ntk}, defined with the other metrics in
Appendix~\ref{app:ntk}), which \citet{htang2024mitigating} identify as the
mechanism behind plasticity loss.

\paragraph{PBT-PPO} \citep{jaderberg2017population} trains a population of
$N$ PPO learners, each with its own networks, optimiser state and
environments.  The members take turns, one update at a time, so the population
spends the same number of environment steps as a single PPO run and meets
every task switch at the same step.  Every $10$ updates of each member, the
members are ranked by their mean training return since the previous exploit
step, and each member in the bottom $20\%$ (at least one) copies the weights of
a member drawn uniformly from the top $20\%$ and restarts its Adam state. 
The
schedule is fixed in updates, so exploit steps do not coincide with task
switches.  The elite is the member with the highest training return and, the centroid is the mean of the members' weights.  We tried $N = 2$ and
$N = 8$ and report, in each figure, the one with the higher cumulative return:
$N = 8$ in every continual setting, and $N = 2$ on some of the stationary and
two-task settings.

\subsubsection{Hyperparameters of every method and environment}
\begin{table}[!htbp]
\centering
\setlength{\belowcaptionskip}{6pt}
\caption{Hyperparameters for NE. $P$: candidates per generation (the
re-scored archive included). $R$: episodes per candidate. $G$: generations per
task. ES: fitness shaping (z-score or centred ranks), optimiser, width
$\sigma$ and learning rate $\alpha$. GA: mutation width $\sigma$ and archive
fraction $\rho = \mu/P$. GA + Novelty uses the GA's $\sigma$, $\rho = 0.5$ and
$k = 3$, and is run only on the classic-control environments and DeepSea. On
MiniGrid the GA uses $P = 256$ and $R = 6$, the same budget. On MountainCar and
Kinetix the GA is the variant described above, and $\sigma$ is its initial
width.}
\label{tab:hyper_ne}
\small
\setlength{\tabcolsep}{4pt}
\begin{tabular}{lrrrllrrrr}
\toprule
 & \multicolumn{3}{c}{shared} & \multicolumn{4}{c}{ES} &
   \multicolumn{2}{c}{GA} \\
\cmidrule(lr){2-4}\cmidrule(lr){5-8}\cmidrule(lr){9-10}
Environment & $P$ & $R$ & $G$ & shaping & optimiser & $\sigma$ & $\alpha$ &
       $\sigma$ & $\rho$ \\
\midrule
CartPole-v1    & $512$ & $3$ & $200$ & z-score & SGD  & $0.1$  & $0.05$  & $0.5$  & $0.5$ \\
Acrobot-v1     & $512$ & $3$ & $200$ & z-score & SGD  & $0.1$  & $0.05$  & $0.5$  & $0.5$ \\
MountainCar-v0 & $512$ & $3$ & $200$ & z-score & SGD  & $0.1$  & $0.05$  & $0.5$  & $0.5$ \\
DeepSea        & $512$ & $3$ & $200$ & z-score & SGD  & $0.1$  & $0.05$  & $0.5$  & $0.5$ \\
MiniGrid       & $512$ & $3$ & $200$ & z-score & SGD  & $0.1$  & $0.05$  & $0.01$ & $0.1$ \\
HalfCheetah    & $512$ & $3$ & $16$  & z-score & SGD  & $0.04$ & $0.005$ & $0.01$ & $0.1$ \\
Kinetix        & $512$ & $1$ & $200$ & ranks   & Adam & $0.02$ & $0.05$ & $0.5$ & $0.5$ \\
\bottomrule
\end{tabular}
\end{table}

\begin{table}[!htbp]
\centering
\setlength{\belowcaptionskip}{6pt}
\caption{Hyperparameters for RL, shared by PPO and its four continual
variants. $N$: parallel environments. $L$: rollout length. $K$: epochs per
update. $M$: minibatches per epoch. $U$: updates per task. Value coefficient
$0.5$ and entropy coefficient $0.01$ in every environment. The settings
specific to each variant are in Table~\ref{tab:hyper_rl_variants}.}
\label{tab:hyper_rl}
\small
\setlength{\tabcolsep}{4pt}
\begin{tabular}{lrrrrrrrrr}
\toprule
Environment & $N$ & $L$ & $K$ & $M$ & $\gamma$ & GAE $\lambda$ & $\alpha$ & clip $\varepsilon$ & $U$ \\
\midrule
CartPole-v1    & $2048$ & $50$ & $10$ & $32$ & $0.95$  & $0.95$ & $3{\times}10^{-4}$ & $0.2$ & $1500$ \\
Acrobot-v1     & $2048$ & $50$ & $10$ & $32$ & $0.99$  & $0.95$ & $1{\times}10^{-4}$ & $0.2$ & $1500$ \\
MountainCar-v0 & $2048$ & $50$ & $10$ & $32$ & $0.99$  & $0.95$ & $3{\times}10^{-4}$ & $0.2$ & $1500$ \\
DeepSea        & $2048$ & $50$ & $10$ & $32$ & $0.99$  & $0.95$ & $3{\times}10^{-4}$ & $0.2$ & $36$ \\
MiniGrid       & $2048$ & $50$ & $4$  & $16$ & $0.99$  & $0.95$ & $5{\times}10^{-4}$ & $0.2$ & $3072$ \\
HalfCheetah    & $512$  & $20$ & $10$ & $32$ & $0.97$  & $0.95$ & $3{\times}10^{-4}$ & $0.3$ & $2400$ \\
HalfCheetah, friction & $4096$ & $20$ & $8$ & $32$ & $0.97$ & $0.95$ & $1{\times}10^{-4}$ & $0.3$ & $300$ \\
Kinetix        & $128$  & $64$ & $8$  & $32$ & $0.995$ & $0.9$ & $5{\times}10^{-5}$ & $0.2$ & $1600$ \\
\bottomrule
\end{tabular}
\end{table}

\begin{table}[!htbp]
\centering
\setlength{\belowcaptionskip}{6pt}
\caption{Hyperparameters specific to the continual variants of PPO, on top of
Table~\ref{tab:hyper_rl}. Classic control: CartPole, Acrobot and MountainCar.
C-CHAIN follows the reference implementation of \citet{htang2024mitigating}:
its categorical settings on the discrete-action environments and its
continuous-control settings on HalfCheetah. No variant is told when the task
changes: the C-CHAIN coefficient, the ReDo schedule and the PBT-PPO exploit
schedule all run unchanged across task switches. PBT-PPO's $N$ is the one
with the higher cumulative return in each figure.}
\label{tab:hyper_rl_variants}
\small
\setlength{\tabcolsep}{4pt}
\begin{tabular}{llll}
\toprule
Variant & Hyperparameter & Value & Environments \\
\midrule
TRAC-PPO & tuner discounts & $0.9, \dots, 0.999999$ (six) & all \\
         & initial scale $S_{0}$ & $\approx 0$ & all \\
\midrule
ReDo-PPO & dormancy threshold $\tau$ & $0.025$ & all \\
         & interval $T_{\mathrm{redo}}$ (updates) & $50$ & classic control \\
         &                                          & $1000$ & other environments \\
         & networks reset & policy and value & all \\
\midrule
C-CHAIN  & target ratio $\kappa$ & $10^{4}$ & discrete actions \\
         &                       & $0.05$   & HalfCheetah \\
         & floor $\lambda_{\min}$ & $1$ & discrete actions \\
         &                        & $0$ & HalfCheetah \\
         & coefficient before warm-up & $1$ & discrete actions \\
         &                            & $100$ & HalfCheetah \\
         & warm-up (updates) & $10$ & discrete actions \\
         &                   & $50$ & HalfCheetah \\
         & averaging window (updates) & $50$ & discrete actions \\
         &                            & $100$ & HalfCheetah \\
\midrule
PBT-PPO  & population $N$ & $8$ or $2$ & all \\
         & exploit interval (updates) & $10$ & all \\
         & exploit fraction & $0.2$ & all \\
         & initial $\alpha$ & $10^{-4}$ & MountainCar \\
\bottomrule
\end{tabular}
\end{table}

Tables~\ref{tab:hyper_ne} and~\ref{tab:hyper_rl} give the settings per environment.
Within an environment, all NE methods share $P$, $R$ and $G$, and all RL methods share
$N$, $L$ and $U$.  We tuned every method to work well on the stationary version
of each environment.  The continual variants of PPO use the PPO settings of
Table~\ref{tab:hyper_rl} and otherwise follow their original implementations
closely.

\FloatBarrier

\subsection{Metrics}
\label{app:metrics}
This section defines the metrics of the paper other than the neighborhood measures, which Appendix~\ref{app:neighborhood} defines. All of them are measured on the reported agent, the centroid (Section~\ref{sec:eval_criteria}): the mean of the search distribution for ES, the mean of the members' weights for the GA, GA + Novelty and PBT-PPO, and the policy itself for PPO, TRAC-PPO, ReDo-PPO and C-CHAIN.

\paragraph{Return}
The return of an agent on a task is its mean undiscounted return over $10$ episodes with greedy actions, on evaluation seeds that are not used for training. This evaluation is not part of the compute budget, and for NE it is separate from the fitness evaluations that selection uses.

\paragraph{Learning accuracy, forgetting and zero-shot transfer}
Let $\theta_j$ be the centroid saved at the end of task $j$ and $R_{i,j}$ its return on task $i$, for a run of $T$ tasks. Then
\begin{equation}
  \mathrm{LA} = \frac{1}{T}\sum_{i=1}^{T} R_{i,i},
  \qquad
  \mathrm{ZT} = \frac{1}{T-1}\sum_{i=1}^{T-1} R_{i+1,i}.
  \label{eq:la_zt}
\end{equation}
When the tasks form a sequence, forgetting is the return that the final agent has lost on every earlier task \citep{lopez-paz2017},
\begin{equation}
  \mathrm{F} = \frac{1}{T-1}\sum_{i=1}^{T-1} \big(R_{i,i} - R_{i,T}\big).
  \label{eq:forgetting}
\end{equation}
When two tasks alternate, the final agent was always trained on the same one of them, so Eq.~\eqref{eq:forgetting} would measure forgetting of the other task only. We instead measure the return lost on a task at the next switch, $R_{i,i} - R_{i,i+1}$, averaged over every switch and so over both directions.

\paragraph{Cumulative return}
Cum. is the area under the curve of the centroid's return on the current task over the whole run. The x-axis is the environment steps of one NE generation, so that NE, which logs once per generation, and RL, which logs once per update, are integrated on the same clock.

\paragraph{Rescaling}
To put LA and F on one scale across settings, we rescale them per setting,
\begin{equation}
  \widehat{\mathrm{LA}} = \frac{\mathrm{LA} - R_{0}}{R^{\ast} - R_{0}},
  \qquad
  \widehat{\mathrm{F}} = \frac{\mathrm{F}}{R^{\ast} - R_{0}},
\end{equation}
where $R_{0}$ is the return of the untrained network and $R^{\ast}$ the highest mean LA of any method in the setting, as in \citet{agarwal2021deep}. The same affine map is applied to every method, so it changes no comparison within a setting. A policy \emph{solves} a task when its rescaled return is at least $0.5$.

\paragraph{Dormant units}
Following \citet{sokar_dormant_2023}, the score of hidden unit $i$ of layer $\ell$ is its mean absolute activation $h_{i}$ over a batch of probe states $\mathcal{D}$, normalised by the mean over the units of its layer,
\begin{equation}
  \mathrm{score}_{i} = \frac{\mathbb{E}_{s \in \mathcal{D}}\,\lvert h_{i}(s)\rvert}{\frac{1}{H_{\ell}}\sum_{k \in \ell} \mathbb{E}_{s \in \mathcal{D}}\,\lvert h_{k}(s)\rvert},
\end{equation}
where $H_{\ell}$ is the width of the layer. In the $\tanh$ networks (HalfCheetah and Kinetix) the numerator is instead the standard deviation of $h_{i}$ over $\mathcal{D}$, and the denominator the layer mean of the same quantity, as for ReDo-PPO (Appendix~\ref{app:redo}): a $\tanh$ unit typically goes silent by saturating at $\pm 1$, where it is constant and passes no gradient yet has the largest possible mean absolute activation. For ReLU the two scores nearly coincide, since a dead ReLU has zero mean and zero spread. A unit is dormant when $\mathrm{score}_{i} \leq \tau = 0.025$, ReDo's own threshold, and the dormant fraction is the share of dormant units over all hidden layers. $\mathcal{D}$ is $512$ states from rollouts of a uniformly random policy, an equal share from every distinct task of the run, and the same batch scores every checkpoint of a run. A random policy does not read the observation, so every method is scored on the same states. The dormancy age of a unit dormant at the end of task $t$ is the number of consecutive tasks, ending at $t$, at whose end it was dormant.

\paragraph{NTK rank}\label{app:ntk}
The neural tangent kernel \citep{jacot_neural_2018} describes how a change of
the parameters moves the outputs of a network.  For a network $f_{\theta}$
with outputs $f_{\theta}(s) \in \mathbb{R}^{m}$ and probe states
$s_{1}, \dots, s_{n}$, the empirical NTK is the Gram matrix of the output
gradients,
\begin{equation}
  K_{\theta}\big((i, a), (j, b)\big)
  = \nabla_{\theta} f_{\theta, a}(s_{i})^{\top}\,
    \nabla_{\theta} f_{\theta, b}(s_{j}),
  \qquad
  K_{\theta} = J_{\theta} J_{\theta}^{\top},
  \label{eq:ntk}
\end{equation}
where $J_{\theta} \in \mathbb{R}^{nm \times |\theta|}$ stacks the gradient of
every output on every state.  To first order, a parameter change
$\delta\theta$ moves the outputs by $J_{\theta}\,\delta\theta$, and a gradient
step on the loss at state $s_{i}$ moves the output at $s_{j}$ in proportion
to the entry of $K_{\theta}$ that couples them.  Large off-diagonal entries
therefore mean that an update fitted on some states also moves the policy on
others (churn), and a low rank means that the outputs can move in only a few
independent directions, so the network cannot fit new targets that fall
outside them; \citet{htang2024mitigating} observe the rank collapsing
as plasticity is lost.  The same bound applies to NE, since a mutation
$\delta\theta$ also moves the outputs by $J_{\theta}\,\delta\theta$.

We report the effective rank $\exp\big(-\sum_{k} p_{k} \log p_{k}\big)$,
where $p_{k} = \mu_{k} / \sum_{l} \mu_{l}$ and $\mu_{k}$ are the eigenvalues
of $K_{\theta}$, on $n = 128$ probe states of the current task.  On the
discrete-action environments $f_{\theta}$ is the vector of logits centred on their mean
over actions (on Kinetix, per action dimension), which neither the softmax nor
the greedy action depends on, so rescaling the logits leaves the rank
unchanged; on the continuous-action environments it is the action mean.  Where a task
offsets the observations, the offset is folded into the first-layer bias, so
that every network is measured in the coordinates it is trained in.

\paragraph{Weight RMS}
The root mean square of the centroid's weights, $\sqrt{\frac{1}{P}\sum_{k=1}^{P}\theta_{k}^{2}}$, where $P$ is the number of parameters (biases included).

\paragraph{Population diversity}
For the methods that keep a population we measure diversity on $512$ fixed probe states. Behavioural diversity is the fraction of probe states on which the greedy actions of two members differ, averaged over all pairs of members; for continuous actions it is the mean absolute difference between their actions. Genomic diversity is the mean pairwise Euclidean distance between the members' weight vectors divided by $\sqrt{2P}$, so that two members drawn independently from $\mathcal{N}(0,\sigma^2 I)$ are at distance $\sigma$ and the genomic diversity of ES is its search width.

\subsection{Neighborhood analysis}
\label{app:neighborhood}
This section defines the three measures of the neighborhood of a policy in weight space: its width, a cheaper proxy of the width (the action change), and the neighborhood it shares with the next task. Like the metrics of Appendix~\ref{app:metrics}, all three are measured on the centroid.

\paragraph{Neighborhood width}
The weights $\theta$ of the centroid consist of tensors $\theta_{1}, \dots, \theta_{L}$ (weight matrices and biases). Following \citet{lehmanESMoreJust2018a}, we perturb the weights with Gaussian noise. As \citet{li2018visualizing}, we scale the noise to the norm of the weights, because a ReLU network's function does not change when a layer is rescaled \citep{dinh2017sharp}; we do so per tensor. A perturbation of radius $\epsilon$ draws Gaussian noise $\xi_{\ell}$ for every tensor and rescales it to that tensor's norm,
\begin{equation}
  \tilde\theta_{\ell} = \theta_{\ell} + \epsilon\,\lVert\theta_{\ell}\rVert\,\frac{\xi_{\ell}}{\lVert\xi_{\ell}\rVert},
  \qquad \ell = 1, \dots, L.
  \label{eq:perturbation}
\end{equation}
Let $s(\epsilon)$ be the share of $64$ such perturbations whose policy no longer solves the task the centroid was trained on (solving as defined in Appendix~\ref{app:metrics}). The neighborhood width is the radius at which $s(\epsilon) = 1/2$, interpolated log-linearly on the ladder $\epsilon \in \{0.001, 0.003, 0.01, 0.03, 0.1, 0.3, 1, 3\}$. It is measured over the last five tasks of a run, only at checkpoints that solve their own task: around a failed policy lies a plateau of failure, which the measure would read as very wide. Each trial contributes one width, and a method's width relative to PPO on a panel is the ratio of the geometric means over trials. Appendix~\ref{app:basin_measure} shows that the ordering of the methods does not depend on the two thresholds that define the width: the rescaled return at which a task counts as solved ($0.5$) and the share of perturbed policies that must have lost the task ($1/2$). On the tanh networks of HalfCheetah and Kinetix the noise has the same standard deviation for every weight instead (see Appendix~\ref{app:basin_noise_scale} for why).

\paragraph{Action change}
A cheaper proxy of the width, used to follow it over training and on the tanh networks, is how often a perturbation changes the greedy action. On $512$ states from a rollout of a uniformly random policy in the task, it is the fraction of states on which the greedy action of $\tilde\theta$ (Eq.~\ref{eq:perturbation}, $\epsilon = 0.1$, $8$ perturbations) differs from that of $\theta$. A small action change is a wide neighborhood.

\paragraph{Shared neighborhood}
For a centroid $\theta_{t}$ that solves task $t$, the shared neighborhood is the share of $64$ perturbations (Eq.~\ref{eq:perturbation}, $\epsilon = 0.1$) whose policy solves both task $t$ and task $t+1$, averaged over the checkpoints of the second half of the run. Appendix~\ref{app:shared_basin} shows that its correlation with LA $-$ F does not depend on the radius.

\section{Performance without task switches}\label{app:noshifts}

\begin{figure}[!htbp]
    \centering
    \includegraphics[width=0.84\linewidth]{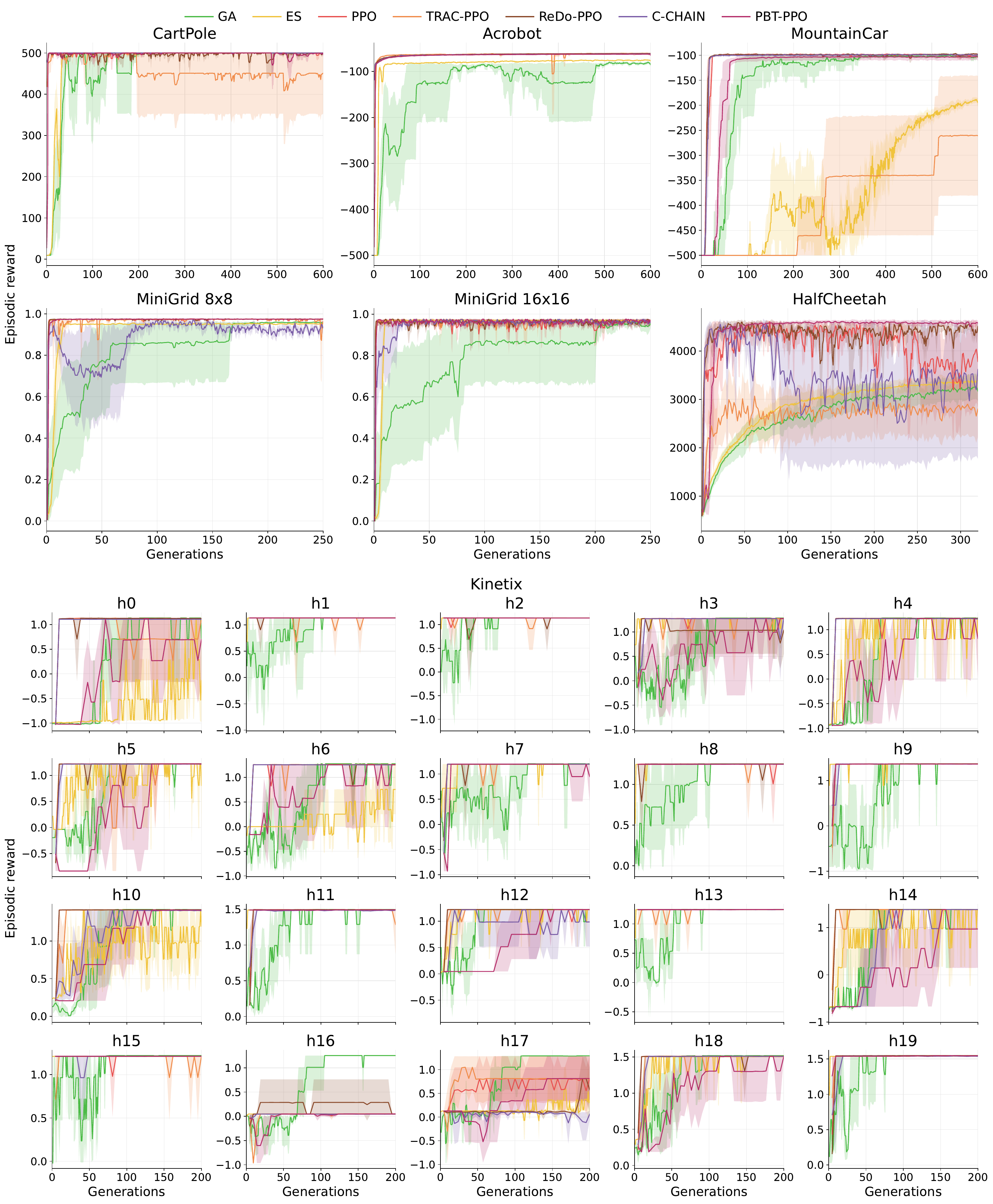}\\[2pt]
    \footnotesize
    \setlength{\tabcolsep}{3pt}
    \begin{tabular}{lrrrrrrr}
    \toprule
    Environment & GA & ES & PPO & TRAC & ReDo & C-CHAIN & PBT \\
    \midrule
    CartPole & 280\,\textcolor{gray}{\scriptsize 34} & 289\,\textcolor{gray}{\scriptsize 26} & 297\,\textcolor{gray}{\scriptsize 2} & 272\,\textcolor{gray}{\scriptsize 3} & 291\,\textcolor{gray}{\scriptsize 3} & 300\,\textcolor{gray}{\scriptsize 3} & 296\,\textcolor{gray}{\scriptsize 3} \\
    Acrobot & -78.1\,\textcolor{gray}{\scriptsize 22} & -52.0\,\textcolor{gray}{\scriptsize 10} & -38.0\,\textcolor{gray}{\scriptsize 225} & -38.7\,\textcolor{gray}{\scriptsize 33} & -38.1\,\textcolor{gray}{\scriptsize 209} & -38.6\,\textcolor{gray}{\scriptsize 70} & -39.7\,\textcolor{gray}{\scriptsize 5} \\
    MountainCar & -90.0\,\textcolor{gray}{\scriptsize 68} & -220\,\textcolor{gray}{\scriptsize 196} & -64.7\,\textcolor{gray}{\scriptsize 13} & -237\,\textcolor{gray}{\scriptsize 239} & -64.3\,\textcolor{gray}{\scriptsize 13} & -64.8\,\textcolor{gray}{\scriptsize 15} & -79.1\,\textcolor{gray}{\scriptsize 44} \\
    MiniGrid 8$\times$8 & 3.78\,\textcolor{gray}{\scriptsize 25} & 3.83\,\textcolor{gray}{\scriptsize 8} & 3.87\,\textcolor{gray}{\scriptsize 2} & 3.86\,\textcolor{gray}{\scriptsize 2} & 3.88\,\textcolor{gray}{\scriptsize 2} & 3.65\,\textcolor{gray}{\scriptsize 2} & 3.89\,\textcolor{gray}{\scriptsize 2} \\
    MiniGrid 16$\times$16 & 3.79\,\textcolor{gray}{\scriptsize 31} & 3.88\,\textcolor{gray}{\scriptsize 8} & 3.82\,\textcolor{gray}{\scriptsize 1} & 3.82\,\textcolor{gray}{\scriptsize 2} & 3.82\,\textcolor{gray}{\scriptsize 2} & 3.81\,\textcolor{gray}{\scriptsize 3} & 3.82\,\textcolor{gray}{\scriptsize 3} \\
    HalfCheetah & 853\,\textcolor{gray}{\scriptsize 234} & 919\,\textcolor{gray}{\scriptsize 185} & 1294\,\textcolor{gray}{\scriptsize 6} & 865\,\textcolor{gray}{\scriptsize 11} & 1379\,\textcolor{gray}{\scriptsize 8} & 1123\,\textcolor{gray}{\scriptsize 4} & 1415$^\dagger$\,\textcolor{gray}{\scriptsize 21} \\
    Kinetix (mean) & 0.177\,\textcolor{gray}{\scriptsize 37} & 0.179\,\textcolor{gray}{\scriptsize 8} & 0.231\,\textcolor{gray}{\scriptsize 10} & 0.228\,\textcolor{gray}{\scriptsize 10} & 0.226\,\textcolor{gray}{\scriptsize 10} & 0.214\,\textcolor{gray}{\scriptsize 10} & 0.168\,\textcolor{gray}{\scriptsize 43} \\
    \bottomrule
    \end{tabular}
    \caption{\textbf{Stationary tasks.} Top: return of the centroid without task switches (MiniGrid: first $250$ generations; bottom: the twenty Kinetix levels). Table: area under the centroid curve (reward $\times$ generations / 1000) and, in grey, the median generation at which the smoothed curve covers 95\% of its rise to its final value; Kinetix is averaged over levels. $^\dagger$PBT-PPO with $N=8$ members.}
    \label{fig:noshifts_all}
\end{figure}

Differences between methods under task switches can be attributed to the switches only if the methods are well tuned without them.
In Figure~\ref{fig:noshifts_all} we present performance in task switches, where we see that this is the case in overall. PPO, ReDo-PPO, C-CHAIN and PBT-PPO collect the most reward on classic control, MiniGrid and HalfCheetah and converge within a few dozen generations, except on Acrobot, where PPO and ReDo-PPO keep improving slowly. ES and TRAC-PPO are slowest on MountainCar, and the GA ends lowest on Acrobot. On Kinetix ES matches the best method on half of the levels, mostly within a few generations, but falls behind on \texttt{h0}, \texttt{h6} and \texttt{h17}; the GA is slower and lower, but it alone collects reward on \texttt{h16}.

\FloatBarrier

\section{Additional results on the continual settings}\label{app:continual}

\subsection{Return over training}\label{app:all_metrics}
\begin{figure}[H]
    \centering
    \includegraphics[width=\linewidth]{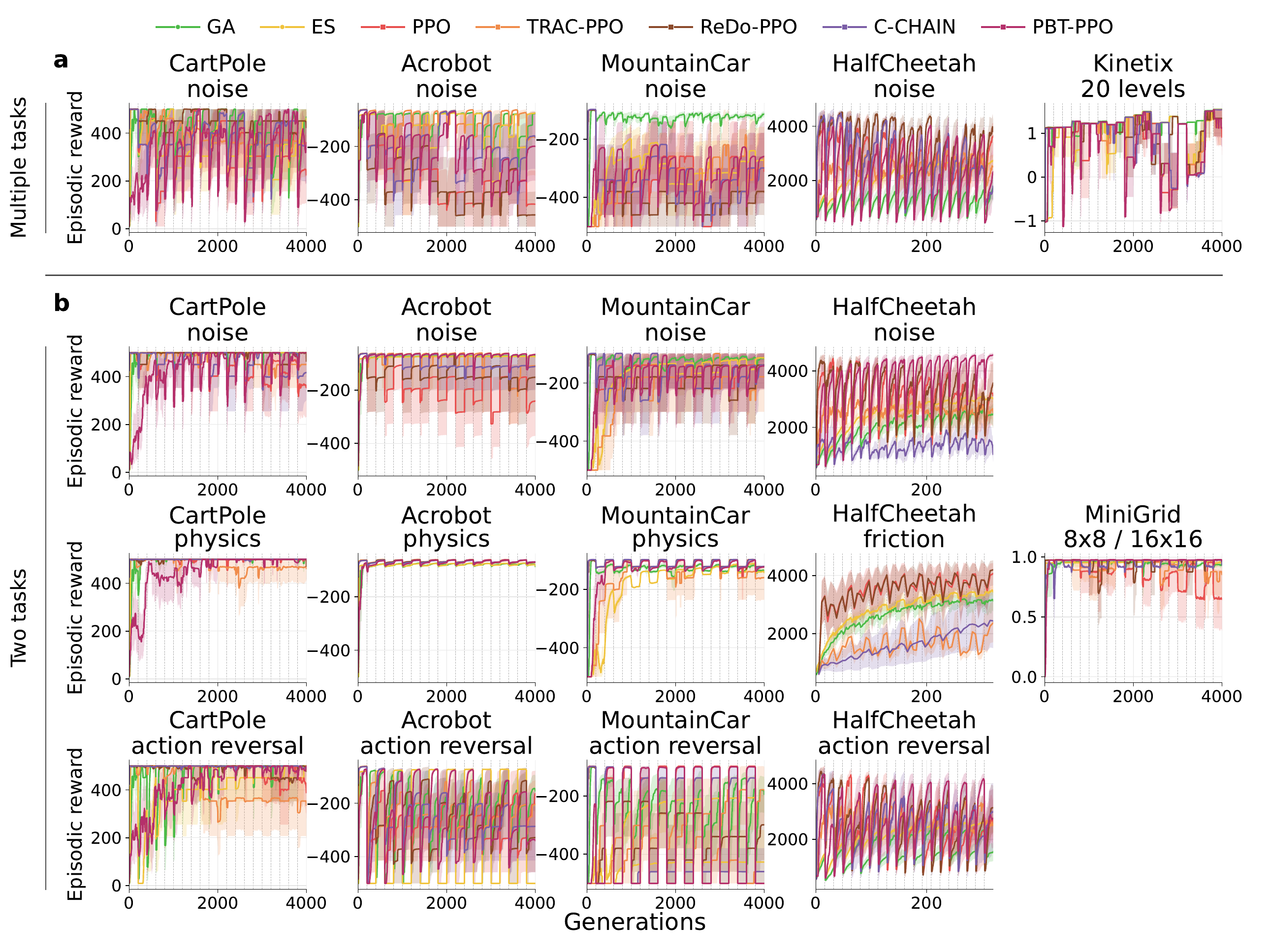}
    \caption{\textbf{Return of the centroid over training.} The settings of Figure~\ref{fig:overall}, in its layout: (a) a sequence of tasks (ten observation offsets, each visited twice; twenty Kinetix levels) and (b) two alternating tasks. Dashed lines are task switches.}
    \label{fig:curves}
\end{figure}

Figure~\ref{fig:curves} shows the return of the centroid over training in every setting of Figure~\ref{fig:overall}, in the same layout. The metrics of Figures~\ref{fig:overall} and~\ref{fig:metrics} are computed from these runs.

\FloatBarrier

\subsection{Symptoms of plasticity loss}\label{app:plasticity_all}
\begin{figure}[H]
    \centering
    \includegraphics[width=0.92\linewidth]{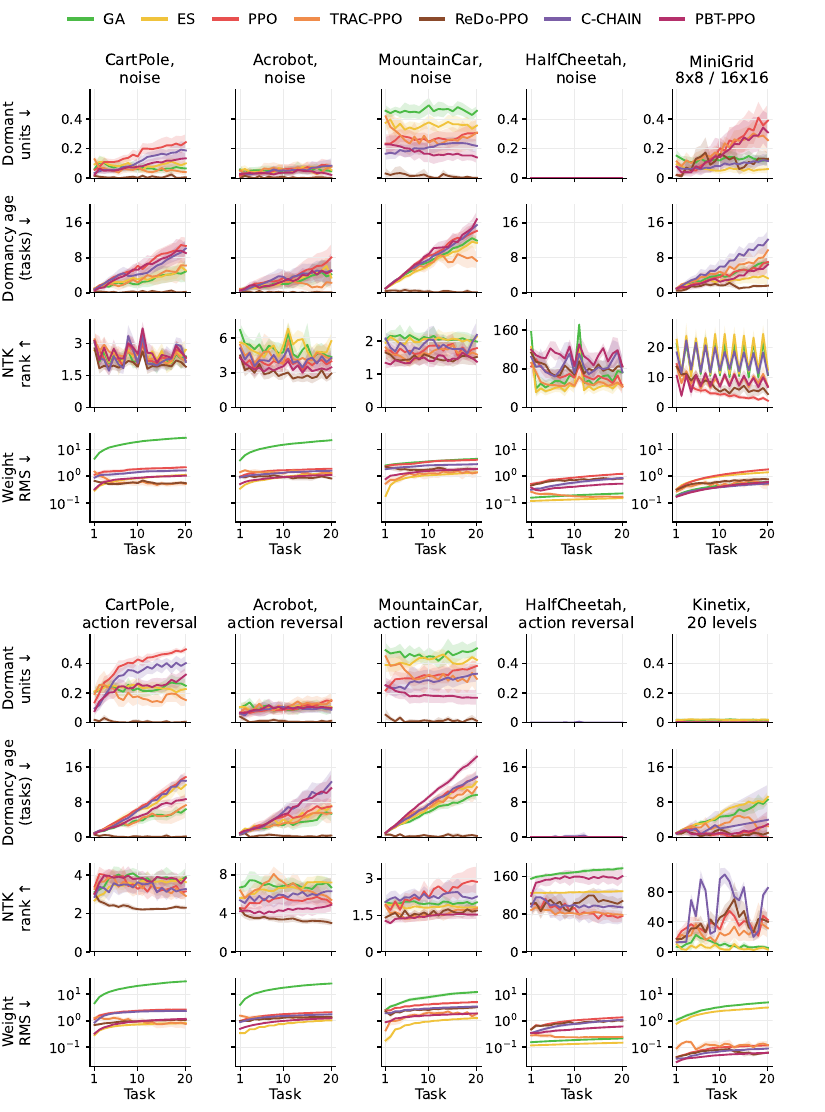}
    \caption{\textbf{Symptoms of plasticity loss in the continual settings.} Figure~\ref{fig:plasticity_symptoms} in ten continual settings (top block: observation noise and MiniGrid; bottom block: action reversal and Kinetix). HalfCheetah and Kinetix have almost no dormant units, so their dormancy lines sit at zero. All rows as defined in Appendix~\ref{app:metrics}.}
    \label{fig:plasticity_all}
\end{figure}

Figure~\ref{fig:plasticity_all} extends Figure~\ref{fig:plasticity_symptoms} to ten continual settings and adds two rows, the dormancy age and the effective rank of the empirical NTK (all rows defined in Appendix~\ref{app:metrics}).

\paragraph{Dormant units.} The dormant fraction of PPO rises over the tasks in every ReLU setting, most on CartPole and MiniGrid, and faster under action reversal than under observation noise. The continual RL methods slow this rise to different degrees, and ReDo-PPO, which resets dormant units, keeps it near zero. The GA and ES stay flat, at a level set by the environment rather than by training: an untrained network of the same shape is as dormant on the same states (Table~\ref{tab:dormant_init}). MountainCar is high for every method because its observations barely vary under a random policy, so many units are silent from the start. HalfCheetah and Kinetix have almost no dormant units. Dormant units also stay dormant longer in RL than in NE, except on MountainCar, where every method but ReDo-PPO keeps them.

\paragraph{NTK rank.} No NE method loses NTK rank. Among the RL methods, PPO and ReDo-PPO lose it on MiniGrid, and ReDo-PPO on CartPole and Acrobot under action reversal. On MiniGrid the rank of every method alternates with the size of the room, because a random walk in the larger room visits fewer distinct views. On HalfCheetah the rank depends mostly on the task, since large observation offsets saturate the tanh units.

\paragraph{Weight RMS.} The GA's weights grow by one to two orders of magnitude in every classic control setting and on Kinetix, but not on HalfCheetah. ES and the RL methods stay close to their initial scale. This growth does not cost the GA its plasticity. Scaling a ReLU layer does not change the greedy action, so selection is blind to the norm and the steps of the centroid are close to a random walk. At these weights the softmax saturates and the policy gradient vanishes, but mutations still change the behaviour, so the GA keeps adapting after every switch.

\begin{table}[H]
\centering
\small
\caption{\textbf{Dormant units of an untrained network against the trained methods, classic control.} Fraction of hidden units below ReDo's threshold on the pooled probe states of Figure~\ref{fig:plasticity_all}: fifty freshly initialised $16\times16$ ReLU policies (mean $\pm$ s.d.), the GA and ES averaged over the twenty tasks, and PPO at the end of the first and of the last task.}
\label{tab:dormant_init}
\setlength{\tabcolsep}{4pt}
\begin{tabular}{lccccc}
\toprule
Setting & Untrained & GA & ES & PPO, task 1 & PPO, task 20 \\
\midrule
CartPole, noise & 0.05 $\pm$ 0.04 & 0.07 & 0.09 & 0.06 & 0.24 \\
CartPole, action reversal & 0.09 $\pm$ 0.05 & 0.24 & 0.24 & 0.14 & 0.50 \\
Acrobot, noise & 0.03 $\pm$ 0.03 & 0.05 & 0.06 & 0.02 & 0.08 \\
Acrobot, action reversal & 0.08 $\pm$ 0.04 & 0.10 & 0.10 & 0.06 & 0.15 \\
MountainCar, noise & 0.42 $\pm$ 0.08 & 0.46 & 0.36 & 0.23 & 0.31 \\
MountainCar, action reversal & 0.49 $\pm$ 0.08 & 0.46 & 0.42 & 0.22 & 0.38 \\
\bottomrule
\end{tabular}
\end{table}

\subsection{GA + Novelty in all seven classic control and DeepSea settings}\label{app:novelty}
\begin{figure}[H]
    \centering
    \includegraphics[width=0.85\linewidth]{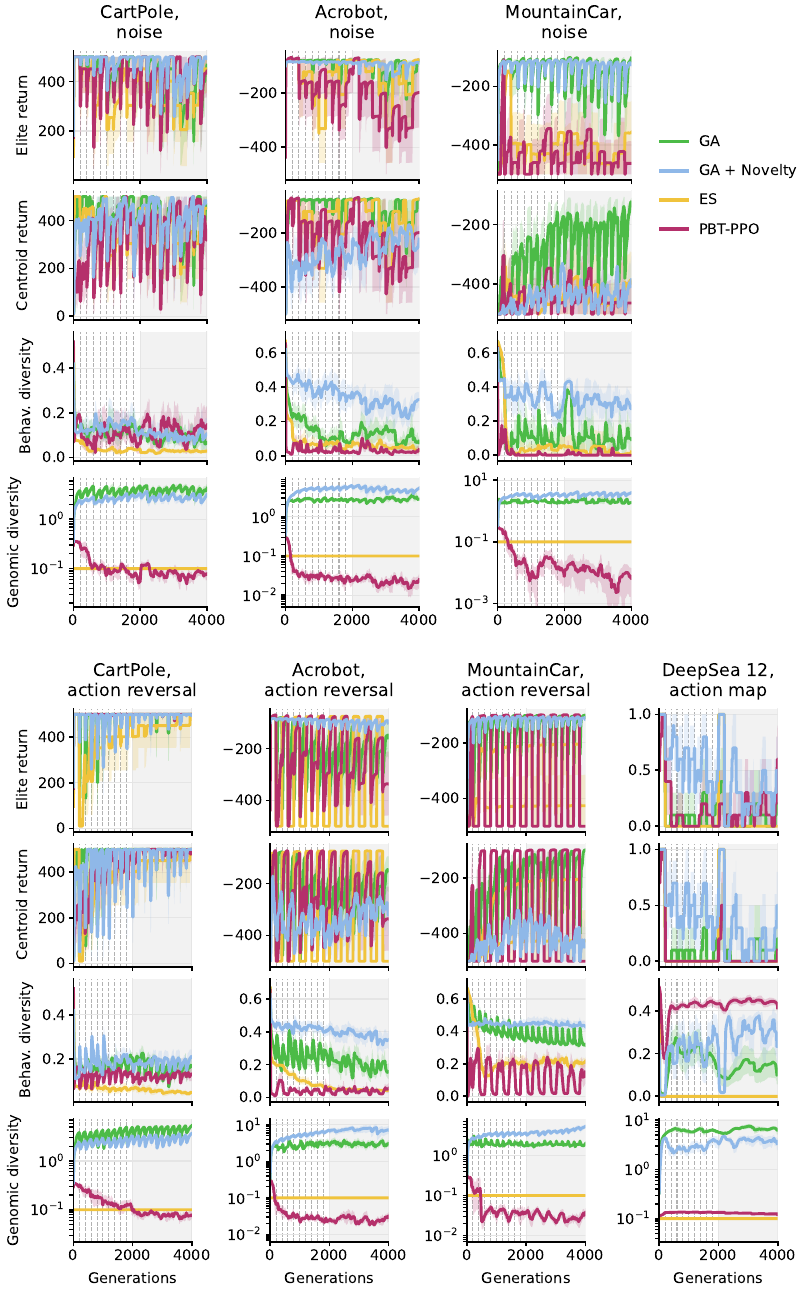}
    \caption{\textbf{Effect of novelty search in all seven settings.} Figure~\ref{fig:novelty} for CartPole, Acrobot and MountainCar under observation noise (top; MountainCar at $\sigma = 0.5$) and under action reversal, and DeepSea (bottom). Rows as in Figure~\ref{fig:novelty}, defined in Appendix~\ref{app:metrics}.}
    \label{fig:novelty_all}
\end{figure}

Figure~\ref{fig:novelty_all} extends Figure~\ref{fig:novelty} from the two settings in which a switch can trap the whole population in a local optimum (MountainCar under noise and DeepSea) to all seven classic control and DeepSea settings. In each setting we compare the run-averaged value of every row between GA + Novelty and each other method (two-sided Mann-Whitney U, Holm over the seven settings); every difference reported below has $p<0.05$.

\paragraph{Behavioural diversity.} GA + Novelty keeps the most behaviourally diverse population on Acrobot and MountainCar under both changes: its members disagree on $0.33$ to $0.45$ of the probe states, against $0.14$ to $0.39$ for the GA, $0.06$ to $0.23$ for ES and $0.01$ to $0.12$ for PBT-PPO. On CartPole it ties the GA under both changes. On DeepSea it is above the GA ($0.25$ against $0.16$) but below PBT-PPO ($0.42$), the only setting in which a method is more behaviourally diverse. Its genomic diversity is $20$ to $170$ times that of PBT-PPO in every setting.

\paragraph{The centroid does not track the elite.} The gap between the elite and the centroid of GA + Novelty is larger than the GA's in every setting, most on Acrobot and MountainCar ($173$ to $293$, against $8$ to $113$ for the GA). The novelty objective spreads the archive over solutions that differ in behaviour, and the average of their weights is none of them. The centroid of GA + Novelty therefore loses cumulative return to the GA's on five settings, CartPole under reversal and Acrobot and MountainCar under both changes, ties on CartPole under noise, and gains only on DeepSea ($0.38$ against $0.13$).

\paragraph{Return against the other methods.} The elite of GA + Novelty gains cumulative return over the GA's on four settings, MountainCar under noise ($+21$) and under reversal ($+31$), Acrobot under reversal ($+116$) and DeepSea ($+0.30$), and ties on the other three. On CartPole the GA's best mutant re-solves the task within a generation of a switch, so there is no collapse for novelty to recover from. It beats ES in all seven settings and PBT-PPO in six, with the largest margins where PBT-PPO is trapped after a switch: on MountainCar ($-146$ against $-433$ under noise, $-133$ against $-312$ under reversal), on Acrobot under reversal ($-101$ against $-231$) and on DeepSea ($0.44$ against $0.21$). The exception is CartPole under reversal, where PBT-PPO's best member returns $484$ against $463$. The centroid does not keep this lead: it beats PBT-PPO's only on CartPole under noise and on DeepSea, and loses to it on Acrobot and MountainCar under reversal. Novelty thus makes the GA's elite the strongest agent in these settings, but not its centroid.

\FloatBarrier

\section{Effect of hyperparameters}\label{app:hyperparameters}

The main text reports one configuration per method (Tables~\ref{tab:hyper_ne} and~\ref{tab:hyper_rl}). Here we vary one hyperparameter at a time and re-run the continual experiment with the same schedule, budget, task draw and seeds, so a difference is attributable to that one value. We look at important hyperparameters for each family: for PPO, the minibatch size, which sets the number of gradient steps per task (Appendix~\ref{app:hyper_ppo}); for the GA and ES, the width of the search distribution (Appendix~\ref{app:hyper_sigma}).

\subsection{PPO: the minibatch size}\label{app:hyper_ppo}
Each PPO update collects a rollout of $NL$ transitions and makes $K$ passes over it, each pass split into $M$ minibatches (Algorithm~\ref{alg:ppo}). We vary the minibatch size $NL/M$ through $M \in \{4, 8, 32, 128\}$, with $32$ the reported value ($\{2, 4, 16, 64\}$ on MiniGrid, reported $16$). On classic control this takes the minibatch from $25{,}600$ transitions down to $800$. Everything else stays fixed, including the data collected, the number of epochs and the learning rate. So a smaller minibatch means proportionally more gradient steps per task ($UKM$), over a $32$-fold range. We run nine settings: CartPole, Acrobot and MountainCar under observation noise and under action reversal, MiniGrid, and HalfCheetah under both perturbations. We report the learning accuracy and forgetting of the centroid (Appendix~\ref{app:metrics}).

\begin{figure}[H]
    \centering
    \includegraphics[width=\linewidth]{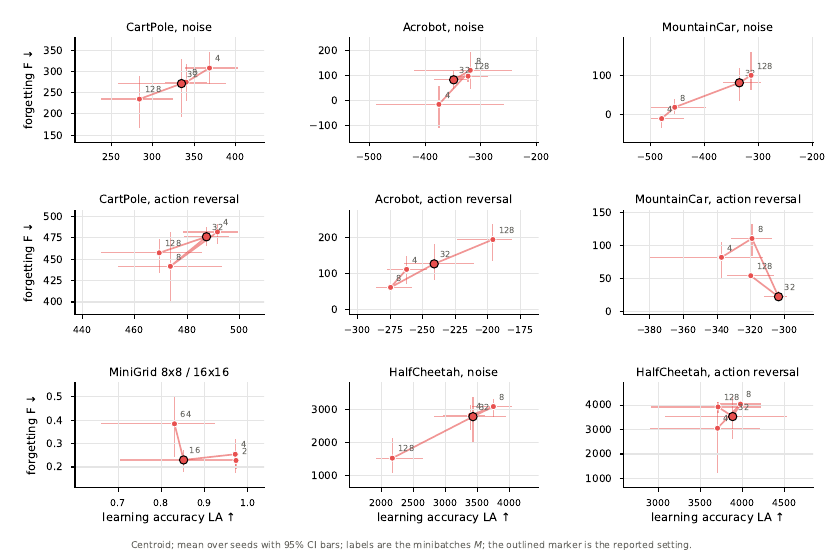}
    \caption{\textbf{Learning accuracy against forgetting as the minibatch size changes.} One panel per setting. The four values of $M$, the number of minibatches per epoch, are joined in order of learning accuracy and labelled. A larger $M$ means a smaller minibatch and more gradient steps per task. The reported value is outlined. Both axes are in the environment's own return units. Lower is more stable and further right is more plastic, so a path along the diagonal trades one for the other.}
    \label{fig:hparam_minibatches_plane}
\end{figure}

\paragraph{The minibatch size moves PPO along its trade-off, not off it.} On seven of the nine panels of Figure~\ref{fig:hparam_minibatches_plane}, learning accuracy and forgetting rise and fall together as $M$ changes, roughly one for one, so what PPO retains (learning accuracy minus forgetting) hardly changes. The two exceptions favour the reported value (MountainCar under action reversal) or come from a few seeds that fail to learn at it (MiniGrid). No value of $M$ is best in every setting on cumulative return, and the reported value is never the worst. The minibatch size therefore decides where PPO sits between plasticity and stability, but not how much it retains. Note that a value that does not learn has nothing to forget: the low forgetting at $M = 4$ on MountainCar under noise, where learning accuracy is near the episode floor, is not evidence of stability.

\subsection{Neuroevolution: the search width of the GA and ES}\label{app:hyper_sigma}

\begin{figure}[H]
    \centering
    \includegraphics[width=0.78\linewidth]{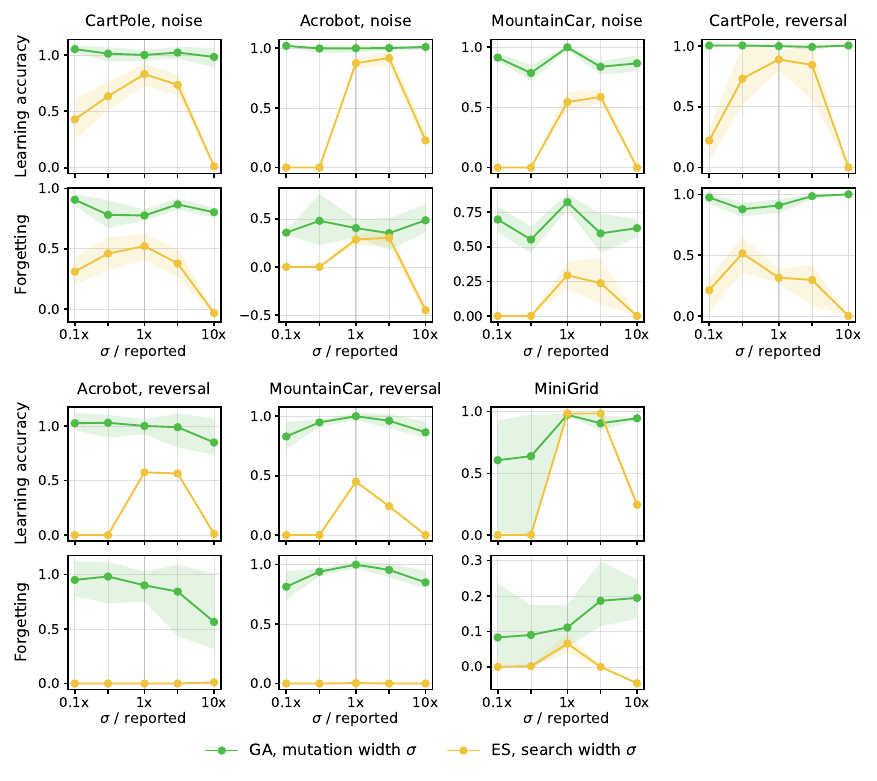}
    \caption{\textbf{Learning accuracy and forgetting of the GA and ES against the search width.} One panel pair per setting: learning accuracy above forgetting, against the multiple of the reported width. Both axes are rescaled so that $0$ is the untrained network and $1$ the best reported method in that setting.}
    \label{fig:hparam_sigma_curves}
\end{figure}

The width of the search distribution is the hyperparameter the GA and ES share, and the one our neighborhood argument rests on. We scale the GA's mutation width ($\sigma = 0.5$) and ES's search width ($\sigma = 0.1$) by $\times 0.1$, $\times 0.3$, $\times 3$ and $\times 10$, with the ES learning rate fixed at $\alpha = 0.05$. We run the three classic control environments under observation noise and under action reversal, and MiniGrid. Figure~\ref{fig:hparam_sigma_curves} shows the learning accuracy and forgetting of the centroid at each width.

\textbf{The GA is robust to its width and ES is not.} The GA learns at every width, and in the classic control environments it is more plastic than ES at every width. ES learns only in a band around the reported value ($\times 1$ and $\times 3$). Narrower, it stays at the untrained score on most panels; ten times wider, it collapses everywhere. This is the band of the toy landscapes (Observation~\ref{obs:toy_rugged}): too narrow a search stops at the first local optimum, and too wide a one reaches the shared peak but does not settle on it. On MiniGrid the GA also loses some seeds at the two narrowest widths.

\textbf{No width reverses a conclusion of the main text.} Wherever ES learns, it forgets less than the GA. Where ES has the best trade-off in Figure~\ref{fig:overall} (CartPole under both kinds of change, and MiniGrid), it keeps more at its good widths than the GA at any width. On Acrobot under noise the two stay tied. Under action reversal on Acrobot and MountainCar ES is frozen on one action ordering at every width at which it learns, while the GA relearns each ordering and forgets the other at every width. An ES width that does not learn has nothing to forget, so its low (or negative) forgetting is not stability.

\section{Effect of the frequency of variation}\label{app:frequency}

\begin{figure}[H]
    \centering
    \includegraphics[width=\linewidth]{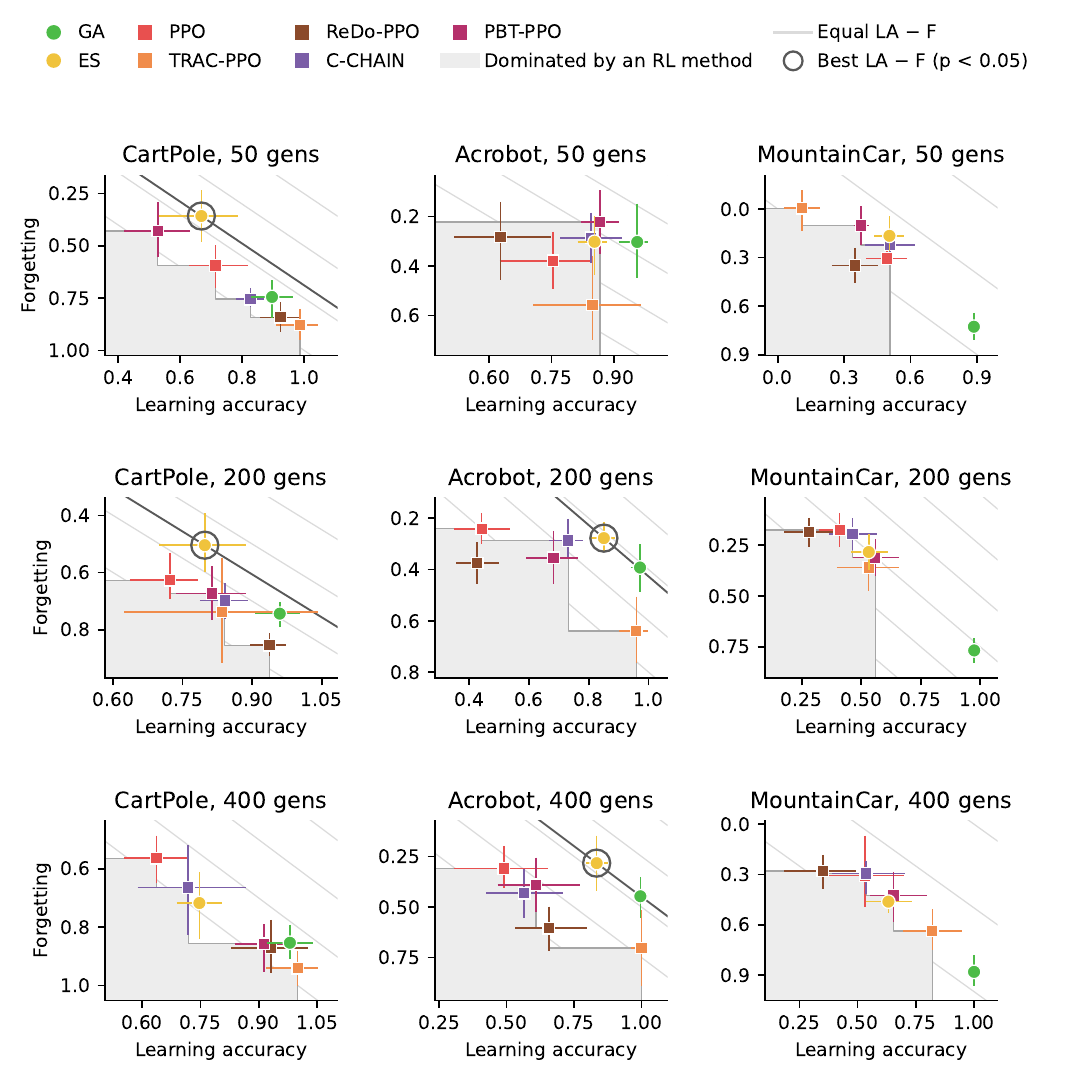}
    \caption{\textbf{Stability-plasticity trade-off against the length of a task, classic control under observation noise.} One row per length, one column per environment; the middle row is the reported setting, as in Figure~\ref{fig:overall}. Learning accuracy against forgetting, the region, lines and ring as in Figure~\ref{fig:overall}, except that both axes are rescaled per environment on one scale shared by the three lengths, so the rows compare directly.}
    \label{fig:frequency}
\end{figure}

Here we explore how the frequency of task switches affects our conclusions. In the main text a task lasts $200$ generations. We shorten it to $50$ and lengthen it to $400$, and keep the budget of $4000$ generations, the tasks, the seeds and the hyperparameters. Figure~\ref{fig:frequency} reports the stability-plasticity trade-off at each length.

\paragraph{The reported length is representative.} At every length and in every environment, a neuroevolution method reaches a trade-off that no RL method matches on both axes: the GA in every panel, with the highest learning accuracy, and ES in most, with lower forgetting. Cumulative return agrees. The GA has the highest return on Acrobot and MountainCar at every length. On CartPole an RL method leads, within the interval of the GA.

\paragraph{Longer tasks cause more forgetting.} Averaged over methods, forgetting grows with the length of a task in every environment. The longer a method trains on one task, the more it specialises to it. Learning accuracy hardly moves, except on MountainCar, whose sparse reward takes many updates to find again after a switch. The RL methods gain the most there, but stay well below the GA.

\section{Population diversity}\label{app:diversity}
Figure~\ref{fig:diversity} tracks three properties of each population over training, defined in Appendix~\ref{app:metrics}: the spread of fitness, behavioural diversity and genomic diversity per weight. It covers the settings on which all four populations log diversity: the six classic control settings of Appendix~\ref{app:continual} and DeepSea 12, the action-map setting of Figure~\ref{fig:novelty}.

Fitness variance rises at every switch for every population, which is the signal selection needs. Behavioural diversity falls within the first task and then stays low, except under GA + Novelty, whose novelty term keeps its members apart: they disagree on $0.25$ to $0.44$ of the probe states in the classic control settings, against $0.08$ to $0.23$ for the GA. In weights, GA + Novelty is the most diverse population in every classic control setting ($4$ to $5.5$ per weight), followed by the GA on CartPole and Acrobot ($2.7$ to $3.9$). ES, whose population is a cloud of width $\sigma$ around the centroid by construction, and PBT-PPO sit near $0.1$. The GA loses this lead on the larger policies: on HalfCheetah and MiniGrid (not shown) its genomic diversity is no larger than that of ES. DeepSea 12 is the one setting where PBT-PPO is the most behaviourally diverse population ($0.43$ of the probe states).

Across settings, a more diverse population learns each task better but forgets it more. When we rank GA, ES and PBT-PPO within each setting, genomic diversity correlates positively with cumulative reward, learning accuracy and forgetting (Spearman $\rho=0.63$, $0.41$ and $0.25$, pooled over the ten settings). The GA carries this ordering. Between ES and PBT-PPO, diversity orders nothing, so the stability of ES does not come from its diversity.

\begin{figure}[H]
    \centering
    \includegraphics[width=\linewidth]{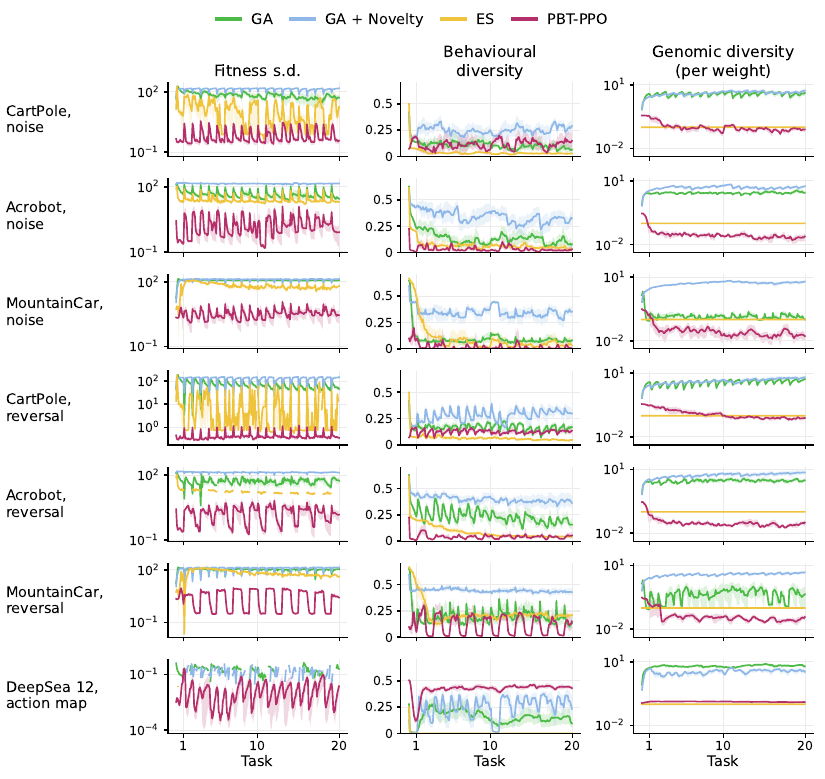}
    \caption{\textbf{Population diversity in the continual settings.} One row per setting, one column per property: standard deviation of fitness, behavioural diversity (disagreement of the members' greedy actions on probe states; mean action distance on HalfCheetah) and genomic diversity per weight. HalfCheetah, MiniGrid and Kinetix are left out: GA + Novelty has no diversity record there.}
    \label{fig:diversity}
\end{figure}
\FloatBarrier

\section{NE in toy landscapes}\label{app:toy}

\newcounter{toyobs}
\newcommand{\toyobservation}[2]{\refstepcounter{toyobs}\label{#1}\paragraph{Observation~\thetoyobs: #2.}}

\paragraph{Setup.} Two tasks, A and B, alternate every $100$ generations. We use two landscapes in two dimensions. In the \emph{smooth} landscape, the peaks of A and B overlap in a plateau of width $h$, and there are no local optima. In the \emph{rugged} landscape, each task is the maximum of four Gaussian peaks: three task-specific peaks and a slightly higher \emph{shared peak} at the origin. We then subtract a ripple of depth $a$ from both tasks. The ripple is zero on a grid that contains every peak, so it lowers no peak, but it turns every grid point into a local optimum. The peaks set where the solutions are; $a$ sets how hard it is to travel between them. Both ingredients are standard: maxima of Gaussians are used in tunable and moving-peaks benchmarks \citep{gallagher2006tunable, branke1999memory}, and the ripple is the cosine term of the Rastrigin function \citep{muhlenbein1991parallel, hansen2009bbob}. Runs start on a peak of A. The \emph{shared region} is where both tasks score within $2\%$ of their maximum: the plateau of the smooth landscape and the top of the shared peak. A seed holds it when its centroid (the ES mean, or the mean of the GA's archive) ends there. We measure the centroid without the ripple, so the shared region is the same at every $a$.

\begin{figure}[!htbp]
    \centering
    \includegraphics[width=\linewidth]{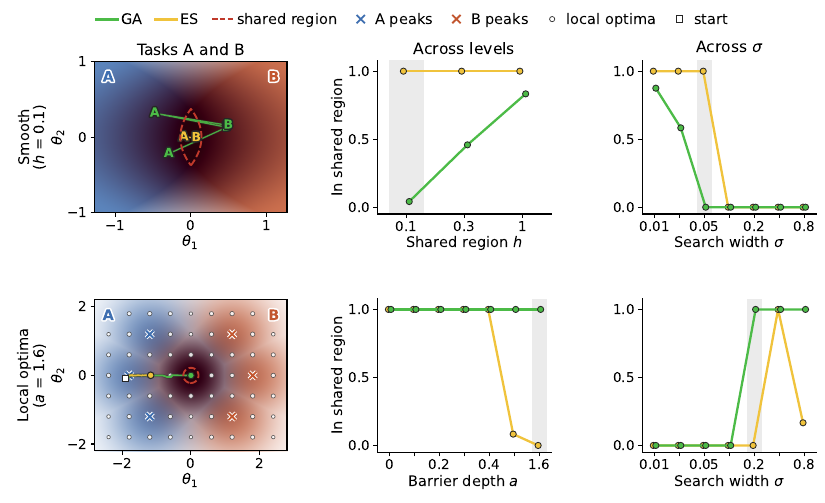}
    \caption{\textbf{Two alternating tasks in toy landscapes.} Top: smooth landscape ($h = 0.1$). Bottom: rugged landscape ($a = 1.6$), with local optima as white dots; at this depth the score between neighbouring dots is zero. Left: task A in blue and task B in red; the shared region is dashed. On the rugged landscape, the centroid path of one seed per method. On the smooth landscape (zoomed), the centroid at the end of each of the last four tasks, marked by the task's letter: the GA switches sides, while ES stays in the shared region. Middle: share of seeds whose centroid ends in the shared region, at every level (grey: the drawn level), with both methods at $\sigma = 0.05$ (smooth) and $\sigma = 0.2$ (rugged). Right: the same share against the search width $\sigma$, at the drawn level (grey: the width used in the other panels).}
    \label{fig:toy_local_optima}
\end{figure}

\toyobservation{obs:toy_smooth}{on the smooth landscape, ES holds the shared region and the GA leaves it}
ES converges to the shared region and stays there at every plateau width $h$ (Figure~\ref{fig:toy_local_optima}, top middle). The GA finds the region too, but leaves it: at the drawn width $h = 0.1$ it ends there in $1$ of $24$ seeds (top left). On the plateau both tasks give every genome the same score, so selection has nothing to hold on to. The GA's archive drifts until a switch pulls it toward the current task's peak. ES has no gradient on the plateau and stays put. This holds only while ES searches narrowly (top right). Once its search is wider than the plateau ($\sigma \geq 0.1$), its samples reach the slope of the current task, and ES follows that slope out of the shared region at every switch. The GA holds the plateau only with a much narrower search ($21$ of $24$ seeds at $\sigma = 0.01$).

\toyobservation{obs:toy_rugged}{on the rugged landscape, the GA finds the shared peak and ES needs an intermediate width}
The depth $a$ of the ripple sets how deep the valleys between neighbouring local optima (the white dots) are. Moving from one dot to the next along one coordinate, the score first falls by $a/2$ and then rises again. At $a = 0$ there are no valleys, and the landscape is a smooth hill up to the shared peak. At the drawn depth $a = 1.6$ a valley ($0.8$) is deeper than the highest peak ($0.70$), so the score between two neighbouring dots is zero. Every local optimum is then an island, and the search must jump from one island to the next to reach the shared peak. The GA does so at every depth (bottom middle): a mutation that lands on a higher island is kept, however unlikely it was. ES moves only where its smoothed score rises, and that score rises toward the shared peak only if its samples reach past the valley. At the width of the middle panel ($\sigma = 0.2$), ES stops at a local optimum once $a \geq 0.8$.

The two methods respond differently to the search width (bottom right). For the GA, wider is better: it holds the shared peak at every $\sigma \geq 0.2$, because selection keeps its archive at the top of the peak whatever the width. ES holds it only at an intermediate width ($\sigma = 0.4$). With a narrower search it cannot see past a valley and stops at the first local optimum. With a wider one ($\sigma = 0.8$) it still reaches the shared peak but ends there in only $4$ of $24$ seeds. Averaged over a wide Gaussian, the shared peak merges with the peaks of the current task, and every switch pulls the centroid away from it. ES cannot hold a target narrower than its search width.

\toyobservation{obs:toy_dims}{more parameters hurt only when they raise the intrinsic dimensionality}
The number of parameters and the intrinsic dimensionality of a landscape can differ. We call the intrinsic dimensionality the number of directions along which the score is curved. Along the other directions the score is flat. A large network can have many parameters but a low intrinsic dimensionality, as has been found when language models are fine-tuned \citep{liangBlessingDimensionality2026}. To separate the two, we fix the number of parameters at $128$ and vary the intrinsic dimensionality $k$ (Table~\ref{tab:toy_dims}). We use the rugged landscape. The first two coordinates carry the peaks as before. The ripple covers $k$ coordinates, including these two, and the other $128 - k$ do not affect the score. Runs start at a grid point of the extra coordinates, so these add no barrier on the way to the shared peak. They only make random steps costly: a step that moves every rippled coordinate pays the ripple on each of them.

With $k = 2$ the landscape has $128$ parameters but the intrinsic dimensionality of the two-dimensional one, and both methods find and hold the shared peak as they do there. Parameters alone therefore do not make the task harder. As $k$ grows, the GA fails. At $a = 0.4$ it holds the shared peak up to $k = 16$ and never from $k = 32$, and a deeper ripple makes it fail sooner. ES holds it up to $k = 128$ unless the ripple is deep. The GA keeps only its best children, and with the cost summed over many coordinates even the best child is worse than its parent. ES needs no single good sample. The cost is shared by all its samples, so it drops out of the ranking, and the ranking still follows the gradient. A larger GA population helps little.

This is a failure to find the shared peak, not to keep it, so it concerns search rather than the stability-plasticity trade-off. It may explain why the GA's advantage fades on larger policies such as HalfCheetah's. That would require their landscapes to have a high intrinsic dimensionality, which we have not shown. When we mutate the saved GA agents of the main experiments, only on HalfCheetah do the children lose clearly more than copies of the parent.

\begin{table}[!htbp]
    \centering
    \small
    \caption{\textbf{Intrinsic dimensionality in the rugged landscape} ($128$ parameters). Seeds out of $24$ whose centroid ends in the shared region, for intrinsic dimensionality $k$ (the number of rippled coordinates) and ripple depth $a$. The other $128 - k$ coordinates do not affect the score. Both methods use $\sigma = 0.4$, their best width among $0.1$, $0.2$ and $0.4$.}
    \label{tab:toy_dims}
    \begin{tabular}{llccccccc}
        \toprule
        & & \multicolumn{7}{c}{intrinsic dimensionality $k$} \\
        \cmidrule(lr){3-9}
        Method & $a$ & $2$ & $4$ & $8$ & $16$ & $32$ & $64$ & $128$ \\
        \midrule
        ES & $0.4$ & $24$ & $24$ & $24$ & $24$ & $24$ & $24$ & $24$ \\
           & $0.8$ & $24$ & $24$ & $24$ & $24$ & $24$ & $24$ & $18$ \\
           & $1.6$ & $24$ & $24$ & $24$ & $24$ & $17$ & $0$  & $0$ \\
        \midrule
        GA & $0.4$ & $24$ & $24$ & $24$ & $24$ & $0$  & $0$  & $0$ \\
           & $0.8$ & $24$ & $24$ & $23$ & $1$  & $0$  & $0$  & $0$ \\
           & $1.6$ & $24$ & $24$ & $4$  & $0$  & $0$  & $0$  & $0$ \\
        \bottomrule
    \end{tabular}
\end{table}

\section{Neighborhood width and return landscapes}\label{app:generalists}

This appendix supports the neighborhood measures defined in Appendix~\ref{app:neighborhood} and draws the return landscapes behind Figure~\ref{fig:overview}.

\subsection{Robustness of the width}\label{app:basin_measure}
\begin{figure}[!htbp]
    \centering
    \includegraphics[width=\linewidth]{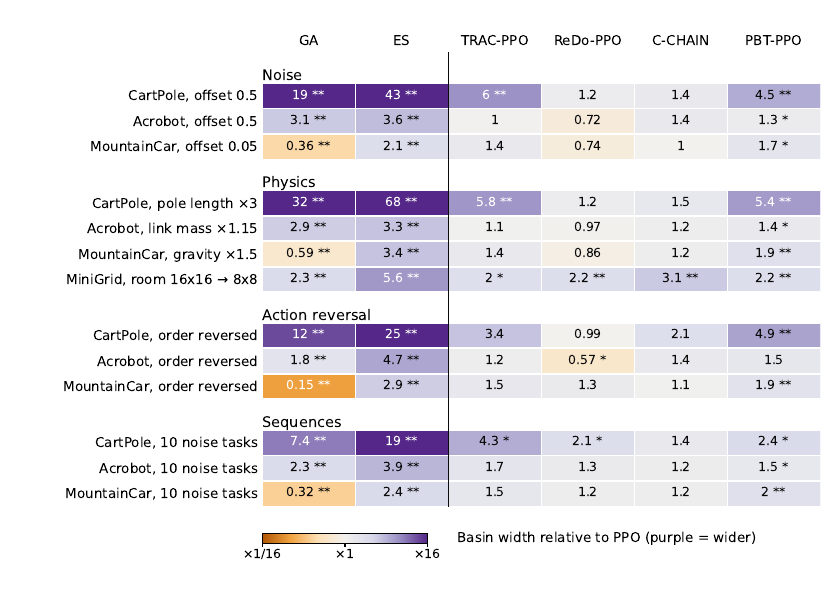}
    \caption{\textbf{Neighborhood width per panel.} Width of the neighborhood around the centroid relative to PPO over the last five tasks, at checkpoints that solve their task, for every method on each classic-control and MiniGrid panel of Figure~\ref{fig:overall}: geometric mean over trials divided by PPO's. Above $1$ is wider than PPO. Stars: Holm-corrected Mann-Whitney U test against PPO over trials ($^{*}$ $p<0.05$, $^{**}$ $p<0.01$).}
    \label{fig:basin_table}
\end{figure}

Figure~\ref{fig:basin_table} gives the width relative to PPO per panel. The rule that only solved checkpoints count (Appendix~\ref{app:neighborhood}) leaves out between $3\%$ (GA) and $27\%$ (PPO) of the checkpoints of the last five tasks.

\begin{figure}[!htbp]
    \centering
    \includegraphics[width=\linewidth]{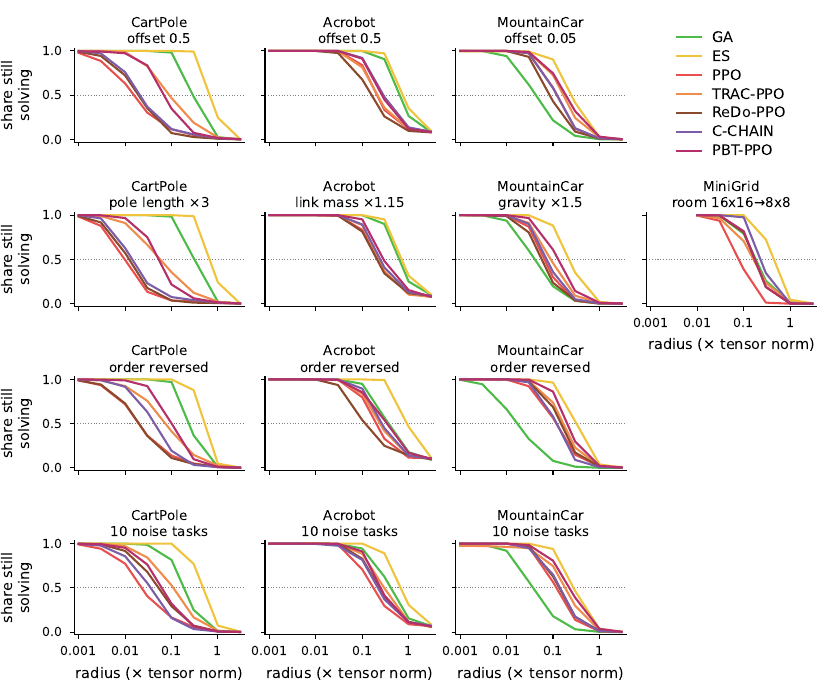}
    \caption{\textbf{The curves the width is read from.} Share of the $64$ random perturbations of the centroid that still solve the task against the radius of the noise, relative to each tensor's norm; mean over the solved checkpoints of the last five tasks, on every classic-control and MiniGrid panel of Figure~\ref{fig:basin}. The width of Figure~\ref{fig:basin} is the radius at which a method's curve crosses the dotted line.}
    \label{fig:basin_curves}
\end{figure}

\paragraph{The thresholds.} Two choices enter the width: the return below which a perturbed policy no longer solves the task ($0.5$), and the share of perturbations that must have crossed it (one half). Table~\ref{tab:basin_levels} re-reads the same curves (Figure~\ref{fig:basin_curves}) under other choices, including the robustness score of \citet{lehmanESMoreJust2018a}, which needs no solved line. Under every definition ES is the widest and ReDo-PPO no wider than PPO, and the per-panel ratios correlate with the reported ones at Spearman $\rho \geq 0.92$. The magnitudes move (ES from $2.2$ to $5.7$ times PPO), which is why Figure~\ref{fig:basin} reports ratios at one level.

\begin{table}[!htbp]
    \centering
    \small
    \caption{\textbf{Neighborhood width under other levels and references.} Width relative to PPO, median over the thirteen classic-control and MiniGrid panels of Figure~\ref{fig:basin}, when the width is read at another point of the same curves: the radius at which the given share of the $64$ perturbed copies has lost the task (\emph{share lost}: rescaled return below $0.5$, or below $0.8$), or the radius at which the median return of the copies relative to the unperturbed policy's own falls below the given level (\emph{median return}), the robustness score of \citet{lehmanESMoreJust2018a}. Stars: Wilcoxon signed-rank test over panels ($^{*}$ $p<0.05$, $^{**}$ $p<0.01$, $^{***}$ $p<0.001$). $\rho$: Spearman correlation of the $78$ per-panel ratios with those of the reported definition ($^\dagger$).}
    \label{tab:basin_levels}
    \setlength{\tabcolsep}{4pt}
\begin{tabular}{lrrrrrrrr}
\toprule
Curve & Level & GA & ES & TRAC-PPO & ReDo-PPO & C-CHAIN & PBT-PPO & $\rho$ \\
\midrule
Share lost, $< 0.5$ & 0.25 & 2.3 & 4.6$^{***}$ & 1.6$^{***}$ & 1.1 & 1.3$^{***}$ & 1.8$^{***}$ & 0.99 \\
 & 0.5$^\dagger$ & 2.3 & 3.9$^{***}$ & 1.5$^{***}$ & 1.2 & 1.4$^{***}$ & 2.0$^{***}$ & -- \\
 & 0.75 & 2.2 & 3.6$^{***}$ & 1.7$^{**}$ & 1.2 & 1.2$^{***}$ & 2.0$^{***}$ & 0.99 \\
 & 0.9 & 1.6 & 2.2$^{***}$ & 1.7$^{**}$ & 1.0 & 1.2$^{*}$ & 1.9$^{**}$ & 0.92 \\
\midrule
Share lost, $< 0.8$ & 0.5 & 2.5 & 4.3$^{***}$ & 1.6$^{**}$ & 1.1 & 1.3$^{***}$ & 1.9$^{***}$ & 0.99 \\
\midrule
Median return & 0.9 & 2.8 & 5.7$^{***}$ & 1.6$^{***}$ & 1.2 & 1.4$^{***}$ & 2.0$^{***}$ & 0.98 \\
 & 0.75 & 2.7 & 3.7$^{***}$ & 1.6$^{***}$ & 1.2 & 1.4$^{***}$ & 2.0$^{***}$ & 0.99 \\
 & 0.5 & 2.4 & 3.4$^{***}$ & 1.6$^{***}$ & 1.2 & 1.4$^{***}$ & 2.1$^{***}$ & 0.99 \\
 & 0.25 & 2.5 & 3.5$^{***}$ & 1.5$^{***}$ & 1.2 & 1.3$^{***}$ & 2.1$^{***}$ & 0.99 \\
\bottomrule
\end{tabular}
\end{table}

\subsection{Significance of neighborhood width}\label{app:basin_test}
The thirteen panels of Figure~\ref{fig:basin} are not independent, since several share an environment. We therefore test within each environment, pooling its panels: a permutation test of the log width against PPO over trials, with the labels permuted within each panel ($4000$ permutations, one-sided). ES is wider than PPO in all four environments (geometric-mean ratio $34$ on CartPole, $3.8$ on Acrobot, $2.7$ on MountainCar and $5.6$ on MiniGrid; each $p<0.001$). The GA is wider in three ($15$, $2.5$ and $2.3$; $p<0.001$) but not on MountainCar ($0.32$). TRAC-PPO is wider in all four ($4.7$, $1.2$, $1.5$ and $2.0$; $p<0.01$), C-CHAIN on CartPole, Acrobot and MiniGrid ($1.3$ to $3.1$) and ReDo-PPO only on CartPole and MiniGrid ($1.3$ and $2.2$).

\paragraph{Why the GA is narrow on MountainCar.} On MountainCar we report the variant of the GA that pulls its archive together whenever the centroid falls behind (Appendix~\ref{app:ga_focus}). In all forty runs its mutation scale sits at its floor of $10^{-5}$ for the second half of training, so the population is a cloud of copies of one network. Selection then never tests the centroid against perturbations near the size at which its neighborhood ends. ES, whose search distribution keeps its width, stays about twice as wide as PPO on the same panels.

\subsection{Neighborhood width over training}\label{app:basin_evolution}
\begin{figure}[!htbp]
    \centering
    \includegraphics[width=\linewidth]{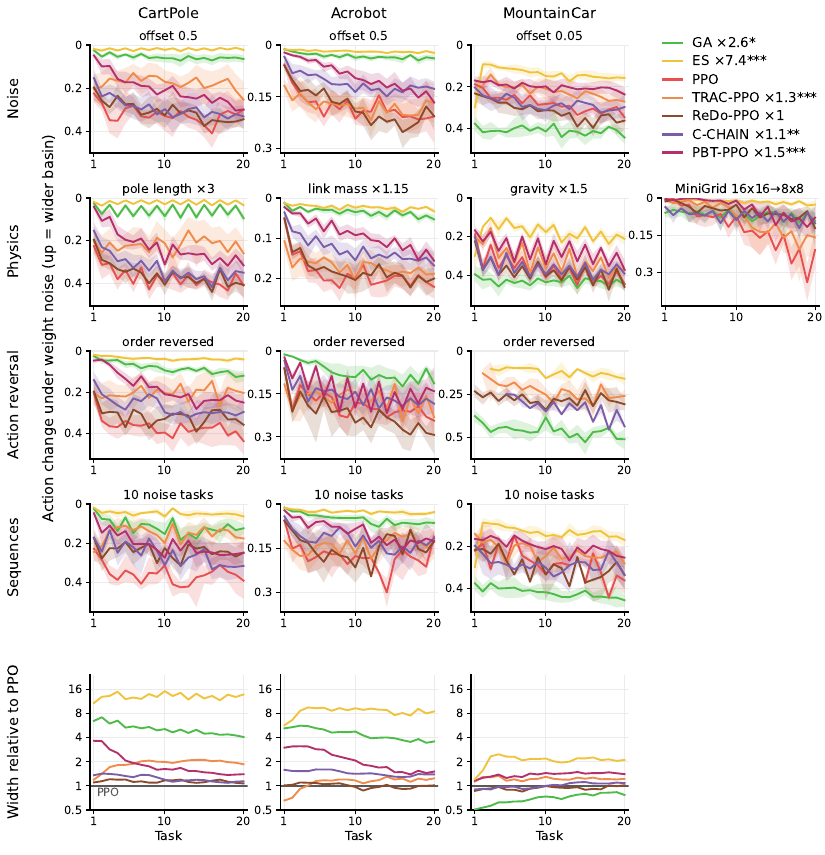}
    \caption{\textbf{Neighborhood width over training, by the action-change proxy.} Top four rows: the change of the greedy action under weight noise of $0.1$ times each tensor's norm, for the centroid saved at the end of every task, on the thirteen ReLU panels of Figure~\ref{fig:basin}, solved checkpoints only. The vertical axis is inverted, so up is a wider neighborhood. Rows are the axis of change, columns the environment. Bottom row: the width relative to PPO, the median over the four panels of an environment, with a running mean over three tasks; MiniGrid has one panel and no line. The legend gives the median width relative to PPO over all thirteen panels by this proxy.}
    \label{fig:basin_evolution}
\end{figure}

Figure~\ref{fig:basin_evolution} follows the width over training by the action-change proxy $\delta$ (Appendix~\ref{app:neighborhood}), which needs no episodes: the width relative to PPO is $\delta_{\mathrm{PPO}}/\delta_{\mathrm{method}}$, and on the thirteen ReLU panels it agrees with the width on the return panel by panel (Spearman $\rho = 0.91$). Every method narrows as training moves it: in the median over the thirteen ReLU panels, the action change over the last five tasks is $1.3$ (TRAC-PPO) to $1.7$ (C-CHAIN) times that over the first three. ES narrows more slowly than PPO, so its lead grows from $6.9$ to $7.4$ times. The GA narrows faster, so its lead falls from $4.9$ to $2.6$ times. The width varies more between environments than between kinds of change, so the bottom row pools the panels of each environment.

\subsection{Neighborhood width on tanh networks}\label{app:basin_noise_scale}
The hidden layers of the HalfCheetah and Kinetix policies are tanh. A tanh network has no rescaling symmetry, so relative noise would only give larger weights larger noise, and the methods' weight scales differ by an order of magnitude in opposite directions (PPO's are eight times those of ES on HalfCheetah and $26$ times smaller on Kinetix). We therefore perturb every weight with the same standard deviation, as \citet{lehmanESMoreJust2018a} did for their tanh Humanoid policies, and measure the action change.

Table~\ref{tab:basin_noise} gives the result; no continual PPO variant is significantly wider than PPO. On Kinetix we also measure the width on the return, following \citet{liangBlessingDimensionality2026} (Table~\ref{tab:basin_return}): there ES is as wide as PPO, and the GA narrower.

\begin{table}[!htbp]
    \centering
    \small
    \caption{\textbf{Neighborhood width relative to PPO on the tanh panels.} Every weight is perturbed by Gaussian noise of standard deviation $0.02$, $0.03$ or $0.1$, the same for every method; last five tasks, every checkpoint. Above $1$ is wider than PPO. Stars: Mann-Whitney U test against PPO over trials, Holm-corrected over the five panels ($^{*}$ $p<0.05$, $^{**}$ $p<0.01$). $0.02$ is the search width of ES on Kinetix and is measured there only. The GA has no run on HalfCheetah under noise with two tasks.}
    \label{tab:basin_noise}
    \begin{tabular}{lcccccc}
        \toprule
        & \multicolumn{3}{c}{ES} & \multicolumn{3}{c}{GA} \\
        \cmidrule(lr){2-4}\cmidrule(lr){5-7}
        Panel & $0.02$ & $0.03$ & $0.1$ & $0.02$ & $0.03$ & $0.1$ \\
        \midrule
        HalfCheetah, offset $0.5$ & -- & $1.48^{**}$ & $1.10$ & -- & -- & -- \\
        HalfCheetah, ground friction & -- & $1.78^{**}$ & $1.35^{**}$ & -- & $1.43^{**}$ & $1.19^{**}$ \\
        HalfCheetah, sign flipped & -- & $1.64^{**}$ & $1.19^{**}$ & -- & $1.28^{**}$ & $1.04$ \\
        HalfCheetah, 10 noise tasks & -- & $1.34^{**}$ & $1.05$ & -- & $1.10$ & $0.96$ \\
        Kinetix, 20 levels & $1.24^{*}$ & $1.30^{*}$ & $1.26^{*}$ & $0.81^{*}$ & $0.85^{*}$ & $1.12$ \\
        \bottomrule
    \end{tabular}
\end{table}

\begin{table}[!htbp]
    \centering
    \small
    \caption{\textbf{Width on the return on Kinetix.} Share of random perturbations of absolute standard deviation $\sigma$ that lose a solved level (score more than $2\%$ of the score range below the unperturbed policy; median over the solved checkpoints at the end of the last two levels, $120$ antithetic directions per checkpoint). Lower is wider. $0.02$ is the search width of ES. Root-mean-square weight in the last column.}
    \label{tab:basin_return}
    \begin{tabular}{lccccc}
        \toprule
        Method & $\sigma = 0.005$ & $0.01$ & $0.02$ & $0.04$ & RMS weight \\
        \midrule
        ES & $0.00$ & $0.00$ & $0.06$ & $0.23$ & $3.1$ \\
        GA & $0.15$ & $0.17$ & $0.28$ & $0.31$ & $4.8$ \\
        PPO & $0.00$ & $0.02$ & $0.07$ & $0.20$ & $0.12$ \\
        \bottomrule
    \end{tabular}
\end{table}

\subsection{The shared neighborhood}\label{app:shared_basin}
Width says how far a policy can move before it loses its task, not whether the next task's solutions lie within that distance. The shared neighborhood (Appendix~\ref{app:neighborhood}) asks the second question at one radius, $\epsilon = 0.1$, the same for every method, so that it describes the landscape and not the method's motion. We probe the gymnax and MiniGrid settings of Figure~\ref{fig:overall}.

\paragraph{The radius.} The correlation with LA $-$ F does not depend on the radius: Spearman $\rho = 0.75$ at $\epsilon = 0.03$, $0.77$ at $0.1$ and $0.73$ at $0.3$, and ES has the largest shared neighborhood in the same five settings at all three. Some settings are easier than others for every method, which alone could produce a correlation over pooled points. A permutation test that shuffles the shared neighborhood among the methods of each setting, and so keeps the differences between settings, gives $p = 5 \times 10^{-5}$ ($20{,}000$ permutations).

\begin{figure}[!htbp]
    \centering
    \includegraphics[width=0.9\linewidth]{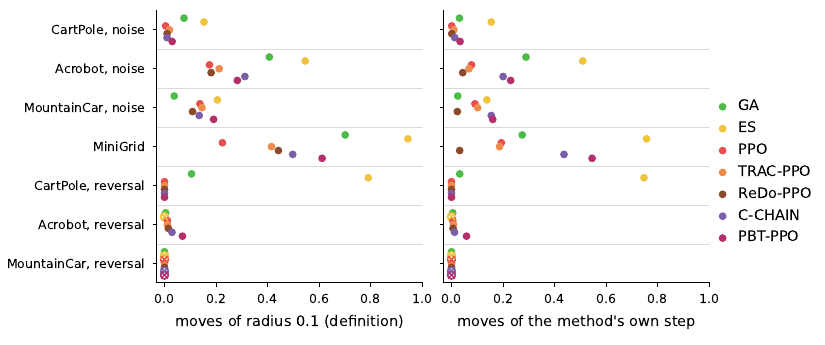}
    \caption{\textbf{The shared neighborhood in every setting.} Share of random moves after which the centroid solves both the task just trained and the next one, in the gymnax and MiniGrid settings of Figure~\ref{fig:overall}. Left: moves of radius $0.1$ of every tensor's norm, the definition, which Figure~\ref{fig:basin} (right) plots against LA $-$ F. Right: moves as large in every tensor as the method's own step over the next task. Crossed: frozen, the method only ever solves one of the two tasks.}
    \label{fig:shared_basin}
\end{figure}

\paragraph{The method's own step.} Forgetting also depends on how far a method moves. Repeating the measure with random moves as large as the method's own step over the next task gives the right panel of Figure~\ref{fig:shared_basin} ($\rho = 0.74$ with LA $-$ F). The GA and ReDo-PPO, which take the largest steps, lose the most: on MiniGrid the GA falls from $0.70$ to $0.27$ and ReDo-PPO from $0.44$ to $0.03$, while ES falls only from $0.94$ to $0.76$.

\begin{figure}[!htbp]
    \centering
    \includegraphics[width=0.55\linewidth]{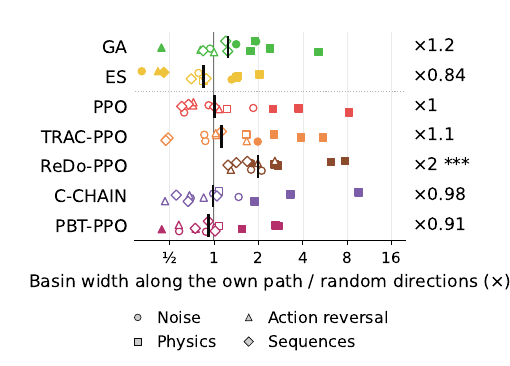}
    \caption{\textbf{Neighborhood width along the method's own path, relative to random directions of the same size.} One point per method and panel of Figure~\ref{fig:basin}: the multiple of the method's step over the next task at which half of the re-scored points on the path no longer solve the task just trained, over the same multiple for random moves as large as the step in every tensor; the bar is the median, a filled point a panel where the paired Wilcoxon test over trials is significant (Holm-corrected), stars a Wilcoxon test over panels. Right of $1$, the path stays inside the neighborhood further than random moves do.}
    \label{fig:basin_direction}
\end{figure}

\paragraph{Along the method's own path.} Random directions are the moves NE makes, but RL follows the gradient, so its neighborhood could be wider along its own path \citep{rahnPolicyOptimizationNoisy2023}. We move each centroid along the step the method actually took over the next task, $\theta_t + f\,(\theta_{t+1} - \theta_t)$ with $f$ from $0.1$ to $8$, and along $64$ random directions of the same size in every tensor, and compare the multiple $f$ at which half of the moves lose the task (Figure~\ref{fig:basin_direction}). For PPO the ratio is $1.00$ in the median over the thirteen panels: its neighborhood is as narrow along its path as in random directions. C-CHAIN ($0.98$) and TRAC-PPO ($1.12$) are the same, and only ReDo-PPO keeps its task further along its path ($1.97$, $p<0.001$). The exception is the three physics changes, where every method's path stays inside the neighborhood far longer, because there the two tasks share their solutions. Random directions are therefore a fair measure of the neighborhood of RL.

\FloatBarrier

\subsection{Return landscapes of the two-task panels}\label{app:landscapes}
\begin{figure}[!htbp]
    \centering
    \includegraphics[width=\linewidth]{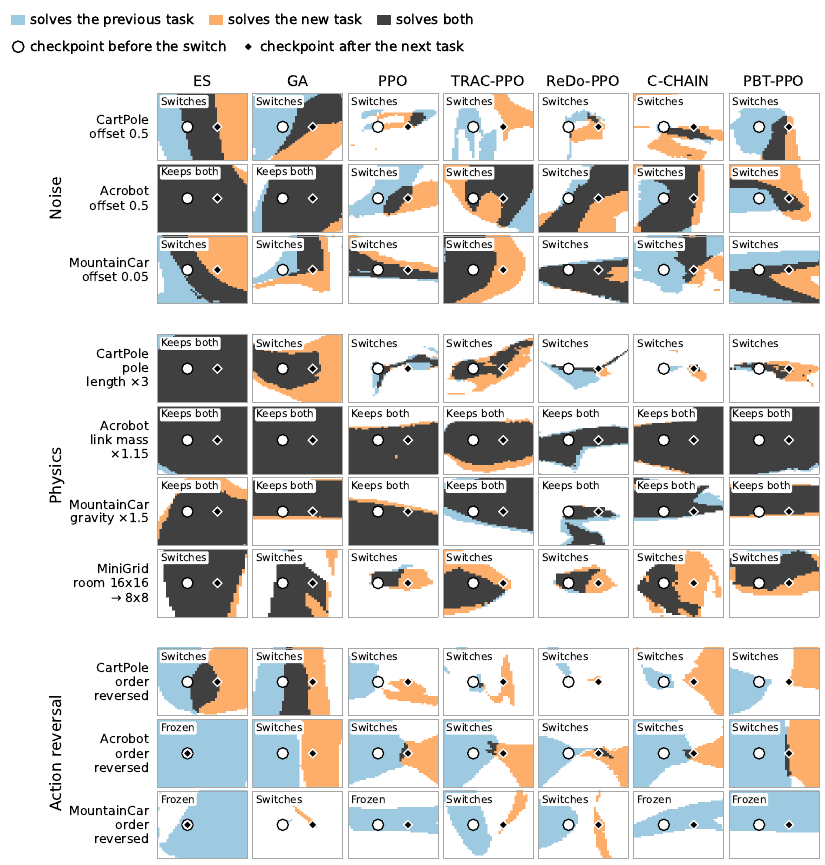}
    \caption{\textbf{Return landscapes around the centroid.} One slice per method (columns) and per two-task setting of Figure~\ref{fig:overall}b with a ReLU network (rows). The circle is the centroid before the switch and the diamond the centroid at the end of the next task; the vertical axis is a random orthogonal direction in units of $0.1$ times the norm of the centroid's weights. Colour marks where a policy solves the previous task (blue), the one just trained (orange) or both (dark), as in Figure~\ref{fig:overview}. Each checkpoint at a switch is classed by its scores on the two tasks: it keeps both (it learns the new task, reaching half the best method's gain over an untrained network, and keeps $90\%$ of its gain on the previous one; filled marker), switches (learns the new task but keeps less; white), is frozen (only ever solves one of the two tasks; crossed) or did not learn (grey). A method takes the class of its mean scores over runs, and the checkpoint shown is, among its checkpoints of that class, the one nearest those mean scores; the corner marks the class. A run that did not move over the task is drawn along its last movement.}
    \label{fig:landscapes}
\end{figure}

Figure~\ref{fig:landscapes} shows a two-dimensional slice of the return landscape through the centroid. The horizontal axis runs from the centroid before a switch to the centroid at the end of the next task, the run's own movement. The vertical axis is a random orthogonal direction relative to the norm of the weights, so we draw slices only for the ReLU networks. At every point we score the policy on both tasks. A checkpoint that keeps both tasks sits on a plateau that spans both colours, one that switches moves from a ridge of one colour to a ridge of the other, and a frozen one does not move.

Figure~\ref{fig:overview} draws the same slice on CartPole under an observation offset. Across all ten settings of Figure~\ref{fig:landscapes} the picture is less clean than on that panel: PPO is stuck at $13\%$ of switches, more than TRAC-PPO ($4\%$) and ReDo-PPO ($9\%$) but not more than C-CHAIN ($11\%$), and under action reversal ES is the method most often stuck.

\FloatBarrier

\section{Analysis of population-based PPO}\label{app:pbt}

PBT-PPO is the RL method with the best LA $-$ F in eight of the eighteen settings, against six for C-CHAIN, three for PPO and one for ReDo-PPO (Figure~\ref{fig:overall}). This appendix asks where that advantage comes from. PBT-PPO trains eight PPO learners that take turns, one update at a time, so the population spends PPO's budget. Every ten updates of each member, the worst fifth copy the weights of a member from the best fifth, and every member keeps PPO's hyperparameters for the whole run (Appendix~\ref{app:methods}). The reported agent is the centroid, the mean of the members' weights. The numbers below rescale each setting between the untrained network ($0$) and the best learning accuracy among the runs compared ($1$). \emph{Kept} is the return on a task after training on the next one.

\paragraph{What does not explain its advantage.} PBT-PPO differs from PPO in five ways besides the number of updates, and none accounts for its advantage. \emph{Width}: it has the widest neighborhood among RL methods ($1.9$ times PPO's, Figure~\ref{fig:basin}), but the width narrows over training (Figure~\ref{fig:basin_evolution}) while its lead in kept return grows over the twenty tasks. \emph{Diversity}: its population is the least behaviourally diverse of the three population methods on six of the eleven tasks of Appendix~\ref{app:diversity}, including those where it keeps the most among RL methods. \emph{Averaging}: the members become near copies within a few tasks, and the best single member is as wide as the mean (in the six noise settings the share of perturbations at relative radius $0.03$ that keep the task differs by at most $0.02$). \emph{Entropy}: its policies stay more stochastic than PPO's ($0.73$ against $0.50$ of the maximum entropy on CartPole under two-task noise), but entropy barely predicts the width of any RL method. \emph{Trainer}: PPO with the full budget on the trainer of PBT-PPO is no better than PPO as reported (Table~\ref{tab:pbt_mechanism}).

\paragraph{Fewer updates per member.} What remains is the budget. Under matched compute each member of PBT-PPO makes an eighth of PPO's updates on a task. A single PPO with the same eighth of the updates, on the trainer, tasks and seeds of PBT-PPO, learns and keeps about as much as PBT-PPO on Acrobot and MountainCar under ten-task noise, and so does PPO with a learning rate ten times smaller (Table~\ref{tab:pbt_mechanism}). Both far exceed PPO as reported, so selection is not needed.

\begin{table}[h]
\centering
\caption{Learning accuracy (LA) and kept return of PPO variants on the ten-task noise settings, rescaled per setting as described at the start of this appendix.}
\label{tab:pbt_mechanism}
\small
\begin{tabular}{lcccccc}
\toprule
 & \multicolumn{2}{c}{Acrobot} & \multicolumn{2}{c}{MountainCar} & \multicolumn{2}{c}{CartPole} \\
 & LA & kept & LA & kept & LA & kept \\
\midrule
PPO (reported) & $0.44$ & $0.29$ & $0.68$ & $0.46$ & $0.79$ & $0.29$ \\
PPO, trainer of PBT-PPO & $0.61$ & $0.48$ & $0.60$ & $0.47$ & $0.65$ & $0.28$ \\
PPO, an eighth of the updates & $1.00$ & $0.76$ & $0.88$ & $0.61$ & $0.93$ & $0.18$ \\
PPO, learning rate $\times 0.1$ & $0.91$ & $0.70$ & $0.78$ & $0.61$ & $1.00$ & $0.19$ \\
PBT-PPO & $0.96$ & $0.65$ & $1.00$ & $0.69$ & $0.96$ & $0.23$ \\
\bottomrule
\end{tabular}
\end{table}

\paragraph{Fewer updates reduce the loss of plasticity.} On Acrobot and MountainCar under noise the learning accuracy of PPO falls over the run, and fewer updates delay that fall. This is why PBT-PPO learns more there.

\paragraph{Fewer updates do not reduce forgetting.} Under ten-task noise PBT-PPO forgets more than PPO on all three classic control tasks, because it learns more (Figure~\ref{fig:overall}). Where a switch needs a solution of both tasks, moving less does not find one. On the three CartPole two-task settings the PPO with an eighth of the updates keeps what PBT-PPO keeps, and both stay far below ES: under action reversal they keep nothing against $0.46$ for ES, under noise $0.40$ and $0.49$ against $0.60$, and under the longer pole $0.67$ and $0.66$ against $1.00$ (see Appendix~\ref{app:shared_basin} for the shared neighborhood of ES).

\paragraph{Fewer updates cost learning on HalfCheetah.} HalfCheetah needs many updates to learn, so the PPO with an eighth of the updates learns less than PBT-PPO (learning accuracy $0.77$ against $0.98$ under two-task noise, $0.58$ against $0.82$ under action reversal); copying the best member makes up the difference. Neither keeps much: $0.15$ and $0.20$ under noise, $0.06$ and $0.09$ under reversal, against $0.39$ and $0.25$ for ES. A PBT-PPO that also perturbs its learning rate and entropy coefficient \citep{jaderberg2017population} keeps more than ES, but we do not report it, since no other RL method tunes its hyperparameters during the run.

\paragraph{Conclusion.} PBT-PPO leads the RL methods because each member makes fewer updates. This reduces the loss of plasticity where PPO loses it, costs learning where a task needs many updates, and does not reduce forgetting.

\appendixreferences

\end{document}